\documentclass[sigconf]{acmart}

\renewcommand\footnotetextcopyrightpermission[1]{} 
\setcopyright{none}

\usepackage{balance}

\def\BibTeX{{\rm B\kern-.05em{\sc i\kern-.025em b}\kern-.08emT\kern-.1667em\lower.7ex\hbox{E}\kern-.125emX}}

\usepackage{subcaption}
\usepackage{tabu}
\usepackage{multirow}
\usepackage{algorithm}
\usepackage{algorithmicx}
\usepackage{algpseudocode}
\usepackage{amssymb}
\usepackage{amsfonts}
\usepackage{mathtools}
\usepackage{enumerate}
\usepackage{enumitem}
\usepackage{amsthm} 
\usepackage{textcomp}
\usepackage{proof}
\usepackage{hyperref}
\usepackage{tikz}
\usepackage{adjustbox}
\usepackage{url}
\usepackage{fixltx2e}
\usepackage{nicefrac}       
\usepackage{microtype}

\usepackage{caption}
\usepackage{booktabs} 
\usepackage{amsmath}
\usepackage{graphicx}
\usepackage{longtable}
\usepackage{tabularx}

\def \figurepath {plots/}
\newcommand{\myparagraph}[1]{\noindent\textbf{#1.}\,}

\DeclareMathOperator*{\argmin}{arg\,min}

\begin{document}

\title{\textsc{Oboe}: Collaborative Filtering for AutoML Model Selection}

\author{
	Chengrun Yang, Yuji Akimoto, Dae Won Kim, Madeleine Udell}
\affiliation{
\institution{Cornell University}}
\email{{cy438,ya242,dk444,udell}@cornell.edu}

%
\begin{abstract}
Algorithm selection and hyperparameter tuning remain two of the most challenging tasks in machine learning.
Automated machine learning (AutoML) seeks to automate these tasks to
enable widespread use of machine learning by non-experts.
This paper introduces \textsc{Oboe}, a collaborative filtering method
for time-constrained model selection and hyperparameter tuning.
\textsc{Oboe} forms a matrix of the cross-validated errors of
a large number of supervised learning models (algorithms together with hyperparameters)
on a large number of datasets,
and fits a low rank model to learn
the low-dimensional feature vectors for the models and datasets
that best predict the cross-validated errors. 
To find promising models for a new dataset,
\textsc{Oboe} runs a set of fast but informative algorithms on the new dataset
and uses their cross-validated errors to infer
the feature vector for the new dataset.
\textsc{Oboe} can find good models under constraints on
the number of models fit or the total time budget.
To this end, this paper develops a new heuristic for active learning in
time-constrained matrix completion based on optimal experiment design.
Our experiments demonstrate that \textsc{Oboe} delivers state-of-the-art performance
faster than competing approaches on a test bed of supervised learning problems.
Moreover, the success of the bilinear model used by \textsc{Oboe}
suggests that AutoML may be simpler than was previously understood.
\end{abstract}

%

%
\maketitle

\section{Introduction}

It is often difficult to find the best algorithm and hyperparameter settings for a new dataset,
even for experts in machine learning or data science.
The large number of machine learning algorithms and
their sensitivity to hyperparameter values make it
practically infeasible to enumerate all configurations.
Automated machine learning (AutoML) seeks to efficiently automate
the selection of model (e.g., \cite{feurer2015efficient, chen2018autostacker, fusi2018probabilistic}) or pipeline (e.g., \cite{drori2018alphad3m}) configurations, and has become more important as
the number of machine learning applications increases.

We propose an algorithmic system, \textsc{Oboe} \footnote{
The eponymous musical instrument
plays the initial note to tune an orchestra.},
that provides an initial tuning for AutoML:
it selects a good algorithm and hyperparameter combination
from a discrete set of options.
The resulting model can be used directly,
or the hyperparameters can be tuned further.
Briefly, \textsc{Oboe} operates as follows.

During an offline training phase, it forms a matrix of the cross-validated errors of
a large number of supervised-learning models (algorithms together with hyperparameters)
on a large number of datasets.
It then fits a low rank model to this matrix to learn
latent low-dimensional meta-features for the models and datasets.
Our optimization procedure ensures these latent meta-features
best predict the cross-validated errors, among all bilinear models.

To find promising models for a new dataset,
\textsc{Oboe} chooses a set of fast but informative models
to run on the new dataset
and uses their cross-validated errors to infer
the latent meta-features of the new dataset.
Given more time, \textsc{Oboe} repeats this procedure using a higher rank
to find higher-dimensional (and more expressive) latent features.
Using a low rank model for the error matrix is a very strong structural prior.

This system addresses two important problems:
1) \textit{Time-constrained initialization}: how to choose a promising initial model under time constraints.
\textsc{Oboe} adapts easily to short times by using a very low rank
and by restricting its experiments to models that will run very fast on the new dataset.
2) \textit{Active learning}: how to improve on the initial guess given further computational resources.
\textsc{Oboe} uses extra time by allowing higher ranks and more expensive
computational experiments, accumulating its knowledge of the new dataset to
produce more accurate (and higher-dimensional) estimates of its latent meta-features.

\textsc{Oboe} uses collaborative filtering for AutoML,
selecting models that have worked well on similar datasets,
as have many previous methods including \cite{bardenet2013collaborative,stern2010collaborative,yogatama2014efficient, feurer2015efficient, misir2017alors, Cunha:2018:CRC:3240323.3240378}.
In collaborative filtering, the critical question is how to characterize dataset similarity
so that training datasets ``similar'' to the test dataset faithfully predict model performance.
One line of work uses dataset meta-features
--- simple, statistical or landmarking metrics ---
to characterize datasets
\cite{pfahringer2000meta,feurer2014using,feurer2015efficient, fusi2018probabilistic, Cunha:2018:CRC:3240323.3240378}.
Other approaches (e.g., \cite{wistuba2015learning}) avoid meta-features.
Our approach builds on both of these lines of work.
\textsc{Oboe} relies on model performance to characterize datasets,
and the low rank representations it learns for each dataset
may be seen (and used) as latent meta-features.
Compared to AutoML systems that compute meta-features of the dataset before
running any models,
the flow of information in \textsc{Oboe} is exactly opposite:
\textsc{Oboe} uses only the performance of various models on the datasets
to compute lower dimensional latent meta-features for models and datasets.

The active learning subproblem is to gain the most information
to guide further model selection.
Some approaches choose a function class to capture
the dependence of model performance on hyperparameters; examples are Gaussian processes \cite{williams2006gaussian,snoek2012practical, bergstra2011algorithms,fusi2018probabilistic, sebastiani2000maximum,herbrich2003fast, mackay1992information, srinivas2009gaussian}, sparse Boolean functions \cite{hazan2018hyperparameter} and decision trees \cite{bartz2004tuning,hutter2011sequential}.
\textsc{Oboe} chooses the set of bilinear models as its function class:
predicted performance is linear in each of the latent model and dataset meta-features.

Bilinearity seems like a rather strong assumption, but confers several advantages.
Computations are fast and easy:
we can find the global minimizer by PCA, and can
infer the latent meta-features for a new dataset using least squares.
Moreover, recent theoretical work suggests that this model class is more general
than it appears: roughly, and under a few mild technical assumptions,
any $m \times n$ matrix with independent rows and columns
whose entries are generated according to a fixed function
(here, the function computed by training the model on the dataset)
has an approximate rank that grows as $\log(m+n)$ \cite{udell2019big}.
Hence large data matrices tend to look low rank.

Originally, the authors conceived of \textsc{Oboe} as a system to
produce a good set of initial models, to be refined by other local search methods,
such as Bayesian optimization.
However, in our experiments, we find that \textsc{Oboe}'s performance,
refined by fitting models of ever higher rank with ever more data,
actually improves faster than competing methods that use local search methods more heavily.

One key component of our system is the prediction of model runtime on new datasets.
Many authors have previously studied algorithm runtime prediction
using a variety dataset features \cite{hutter2014algorithm},
via ridge regression \cite{huang2010predicting},
neural networks \cite{smith2011discovering},
Gaussian processes \cite{hutter2006performance}, and more.
Several measures have been proposed to trade-off
between accuracy and runtime \cite{leite2012selecting,bischl2017mlrmbo}.
We predict algorithm runtime using only the number of samples and features
in the dataset. This model is particularly simple but surprisingly effective.

Classical experiment design (ED) \cite{wald1943efficient,mood1946hotelling,john1975d,pukelsheim1993optimal,boyd2004convex}
selects features to observe to minimize the variance of the parameter estimate,
assuming that features depend on the parameters according to known, linear, functions.
\textsc{Oboe}'s bilinear model fits this paradigm, and so
ED can be used to select informative models.
Budget constraints can be added,
as we do here, to select a small number of promising machine learning models
or a set predicted to finish within a short time budget
\cite{krause2008near,zhang2016flash}.

This paper is organized as follows.
Section~\ref{notation} introduces notation and terminology.
Section~\ref{methodology} describes the main ideas we use in \textsc{Oboe}.
Section~\ref{oboe} presents \textsc{Oboe} in detail.
Section~\ref{experiments} shows experiments.

\section{Notation and Terminology} \label{notation}
\myparagraph{Meta-learning}
Meta-learning is the process of learning across individual datasets or problems,
which are subsystems on which standard learning is performed \cite{lemke2015metalearning}.
Just as standard machine learning must avoid overfitting,
experiments testing AutoML systems must avoid meta-overfitting!
We divide our set of datasets into meta-training, meta-validation and meta-test sets, and report results on the meta-test set.
Each of the three phases in meta-learning --- meta-training, meta-validation and meta-test ---
is a standard learning process that includes training, validation and test.

\begin{figure}
	\begin{subfigure}[t]{0.48\linewidth}
		\includegraphics[width=0.8\linewidth]{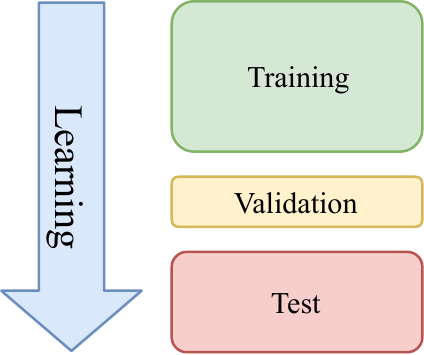}
		\caption{Learning}
	\end{subfigure}
	\begin{subfigure}[t]{0.48\linewidth}
		\includegraphics[width=0.8\linewidth]{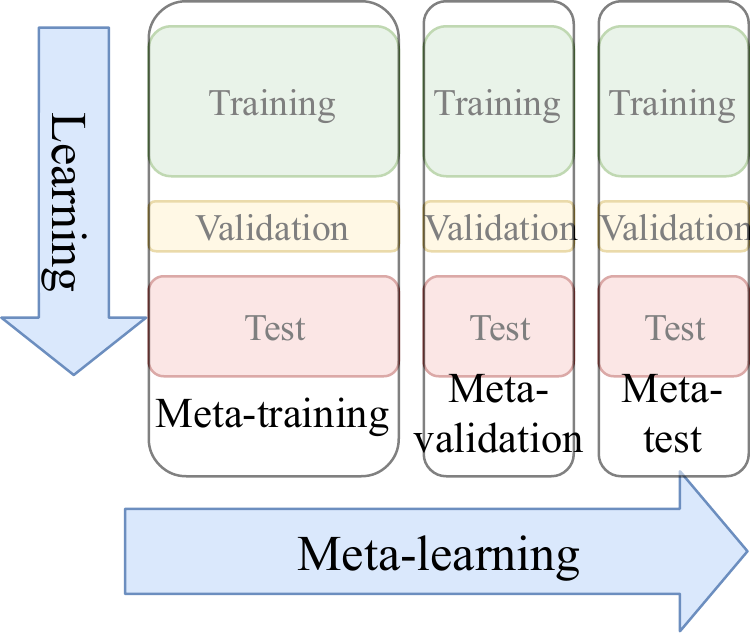}
		\caption{Meta-learning}
	\end{subfigure}
	\caption{Standard vs meta-learning.}
	\label{fig:metalearning}
\end{figure}

\myparagraph{Indexing}
Throughout this paper, all vectors are column vectors. Given a matrix $A \in \mathbb{R}^{m \times n}$,
$A_{i,:}$ and $A_{:,j}$ denote the \textit{i}th row and \textit{j}th column of $A$, respectively. $ i $ is the index over datasets, and $ j $ is the index over models.
We define $[n] = \{1,\ldots,n\}$ for $n \in \mathbb{Z}$.
Given an ordered set $\mathcal{S} = \{ s_1, \ldots, s_k \}$ where $s_1 <  \ldots < s_k \in [n]$,
we write $A_\mathcal{:S} = \begin{bmatrix} A_{:,s_1} & A_{:,s_2} & \cdots & A_{:,s_k} \end{bmatrix}$.

\myparagraph{Algorithm performance}
A \emph{model} $ \mathcal{A} $ is a specific algorithm-hyperparameter combination,
e.g. $ k $-NN with $ k=3 $.
We denote by $\mathcal{A}(\mathcal{D}) $ the expected cross-validation error
of model $ \mathcal{A} $ on dataset $ \mathcal{D} $,
where the expectation is with respect to the cross-validation splits.
We refer to the model in our collection
that achieves minimal error on $ \mathcal{D} $
as the \emph{best model} for $ \mathcal{D} $.
A model $ \mathcal{A} $ is said to be \emph{observed} on $ \mathcal{D} $
if we have calculated $\mathcal{A}(\mathcal{D}) $ by fitting (and cross-validating) the model.
The \emph{performance vector} $ e $ of a dataset $ \mathcal{D} $ concatenates
$\mathcal{A}(\mathcal{D}) $ for each \emph{model} $ \mathcal{A} $ in our collection.

\myparagraph{Meta-features}
We discuss two types of meta-features in this paper.
\emph{Meta-features} refer to metrics used to characterize datasets or models.
For example, the number of data points or
the performance of simple models on a dataset can serve as meta-features of the dataset.
As an example, we list the meta-features used in the AutoML framework auto-sklearn
in Appendix~\ref{supp:metafeature}, Table~\ref{table:metafeatures}.
In constrast to standard meta-features, we use the term
\emph{latent meta-features} to refer to
characterizations learned from matrix factorization.

\myparagraph{Parametric hierarchy}
We distinguish between three kinds of parameters:
\begin{itemize}[noitemsep,nolistsep]
\item \textit{Parameters} of a model (e.g., the splits in a decision tree)
are obtained by training the model.

\item \textit{Hyperparameters} of an algorithm
(e.g., the maximum depth of a decision tree) govern the training procedure.
We use the word \emph{model} to refer to an algorithm together with a particular choice of hyperparameters.

\item \textit{Hyper-hyperparameters} of a meta-learning method
(e.g., the total time budget for \textsc{Oboe})
govern meta-training.
\end{itemize}

\myparagraph{Time target and time budget}
The time target refers to the anticipated time spent running models to infer
latent features of each fixed dimension and can be exceeded.
However, the runtime does not usually deviate much from the target
since our model runtime prediction works well.
The time budget refers to the total time limit for \textsc{Oboe} and is never exceeded.

\myparagraph{Midsize OpenML and UCI datasets} Our experiments use
OpenML \cite{OpenML2013} and UCI \cite{Dua:2017} classification datasets
with between 150 and 10,000 data points and with no missing entries. 

\section{Methodology} \label{methodology}
\subsection{Model Performance Prediction}
It can be difficult to determine \textit{a priori} which meta-features to use
so that algorithms perform similarly well on datasets with similar meta-features.
Also, the computation of meta-features can be expensive (see Appendix~\ref{metafeaturetime}, Figure~\ref{fig:metafeature_calculation}).
To infer model performance on a dataset without any
expensive meta-feature calculations,
we use collaborative filtering to infer latent meta-features for datasets.

\begin{figure}
	\centering
	\includegraphics[width=.9\linewidth]{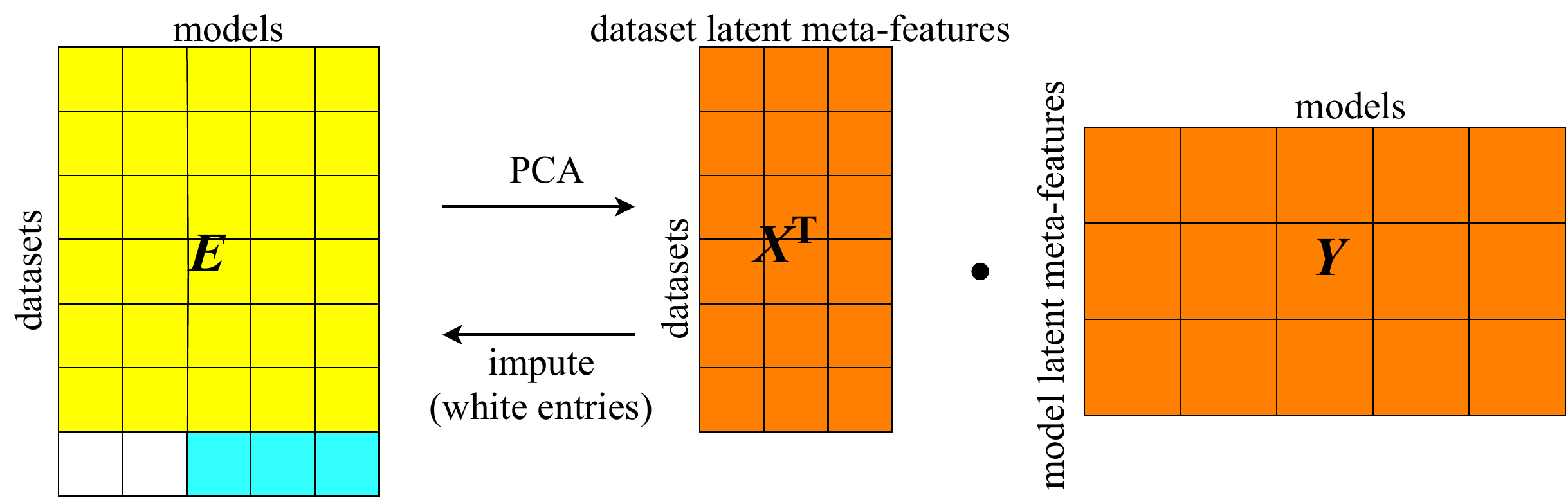}
	\caption{Illustration of model performance prediction via the error matrix $ E $ (yellow blocks only).
		Perform PCA on the error matrix (offline) to compute dataset ($X$) and
		model ($Y$) latent meta-features (orange blocks).
		Given a new dataset (row with white and blue blocks),
		pick a subset of models to observe (blue blocks).
		Use $ Y $ together with the observed models
		to impute the performance of the unobserved models on the new dataset (white blocks).}
	\label{fig:imputation}
\end{figure}

\begin{figure}
\centering
\includegraphics[width=.5\linewidth]{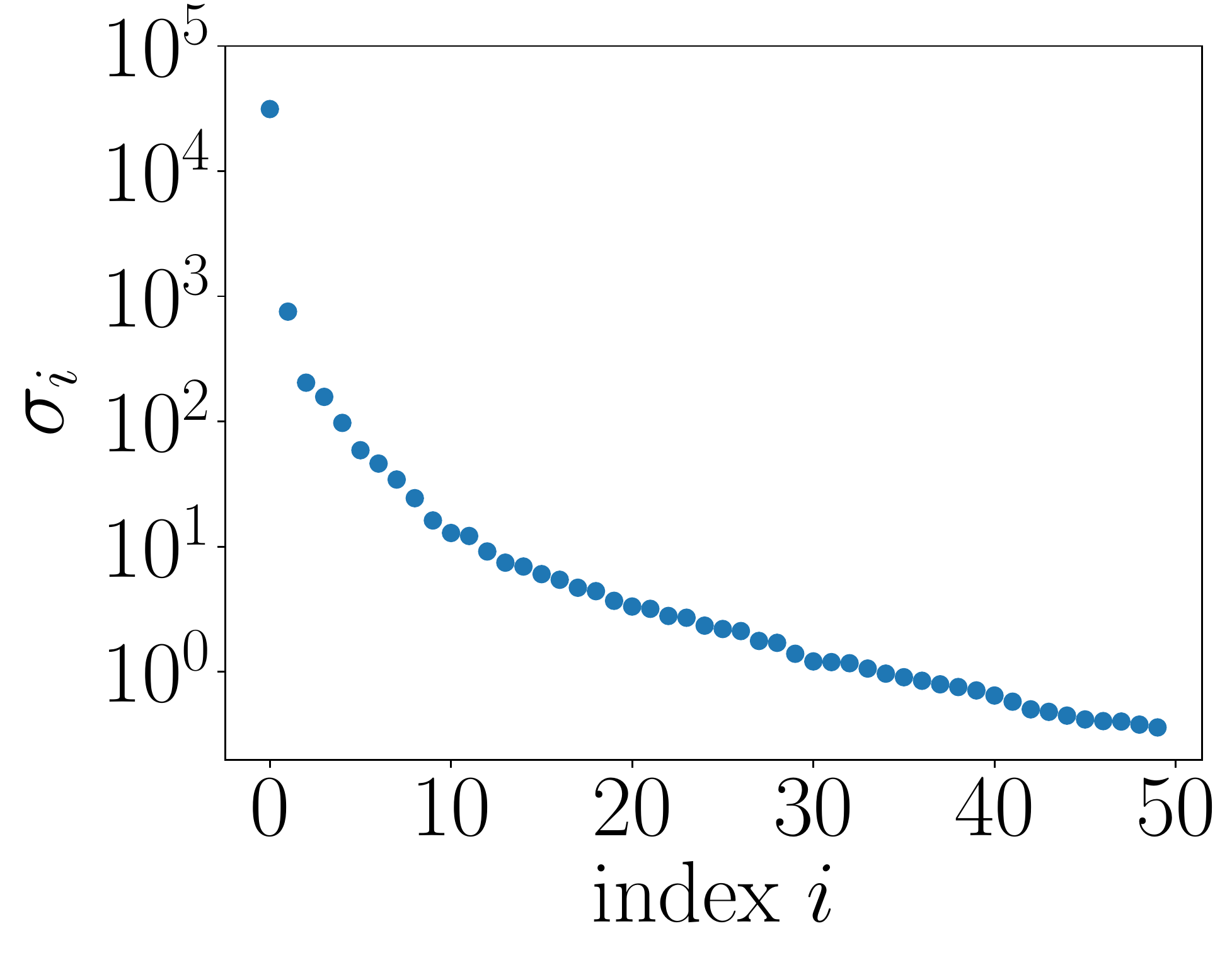}
\caption{Singular value decay of an error matrix. The entries are calculated by 5-fold cross validation of machine models (listed in Appendix~\ref{supp:models}, Table~\ref{table:models}) on midsize OpenML datasets.}
\label{fig:singularValues}
\end{figure}

As shown in Figure~\ref{fig:imputation},
we construct an empirical error matrix $E \in \mathbb{R}^{m \times n}$,
where every entry $E_{ij}$ records the cross-validated error of model $j$ on dataset $i$.
Empirically, $ E $ has approximately low rank: Figure~\ref{fig:singularValues} shows
the singular values $\sigma_i(E)$ decay rapidly as a function of the index $i$.
This observation serves as foundation of our algorithm,
and will be analyzed in greater detail in Section~\ref{section:nmf}.
The value $E_{ij}$ provides a noisy but unbiased estimate
of the true performance of a model on the dataset:
$\mathbb{E} E_{ij} = \mathcal{A}_j(\mathcal{D}_i)$.

To denoise this estimate,
we approximate $E_{ij} \approx x_i^\top y_j$ where $x_i$ and $y_j$ minimize
$ \sum_{i=1}^m \sum_{j=1}^n (E_{ij} - x_i^\top y_j)^2$ with $x_i, y_j \in \mathbb{R}^k$ for $i \in [M]$ and $j \in [N]$; the solution is given by PCA.
Thus $x_i$ and $ y_j $ are the latent meta-features of dataset $i$ and model $j$, respectively.
The rank $k$ controls model fidelity: small $k$s give coarse approximations, while
large $k$s may overfit.
We use a doubling scheme to choose $k$ within time budget; see Section~\ref{s-doubling} for details.

Given a new meta-test dataset, we choose a subset $\mathcal{S} \subseteq [N]$ of models and observe performance $e_j$ of model $j$ for each $j \in \mathcal{S}$.
A good choice of $S$ balances information gain against time needed to run the models;
we discuss how to choose $S$ in Section~\ref{sec:ED}.
We then infer latent meta-features for the new dataset by solving the least squares problem: 
$\text{minimize} \sum_{j \in \mathcal{S}} (e_j - \hat{x}^\top y_j)^2$ with $\hat{x} \in \mathbb{R}^k$.
For all unobserved models,
we predict their performance as $\hat e_j = \hat{x}^\top y_j$ for $j \notin \mathcal{S}$.

\subsection{Runtime Prediction}
Estimating model runtime allows us to trade off between running
slow, informative models and fast, less informative models.
We use a simple method to estimate runtimes, using polynomial regression on
$n^\mathcal{D}$ and $p^\mathcal{D}$, the numbers of data points and features in $\mathcal{D}$,
and their logarithms, since the theoretical complexities of machine learning algorithms we use
are $O \big((n^\mathcal{D})^3, (p^\mathcal{D})^3, (\log(n^\mathcal{D}))^3 \big)$.
Hence we fit an independent polynomial regression model for each model:
\[
f_j = \mbox{argmin}_{f_j \in \mathcal{F}}
\sum_{i=1}^M \left(
f_j(n^{\mathcal{D}_i}, p^{\mathcal{D}_i}, \log(n^{\mathcal{D}_i})) - t_j^{\mathcal{D}_i}
\right)^2,
\,
j \in [n]
\]
where $t_j^\mathcal{D}$ is the runtime of machine learning model $j$ on dataset $\mathcal{D}$,
and $\mathcal{F}$ is the set of all polynomials of order no more than 3.
We denote this procedure by $f_j=$ \texttt{fit\_runtime}$(n, p, t)$.

We observe that this model predicts runtime within a factor of two
for half of the machine learning models on more than 75\% midsize OpenML datasets,
and within a factor of four for nearly all models,
as shown in Section~\ref{section:runtime_prediction}
and visualized in Figure~\ref{fig:runtime_prediction_by_type}.

\subsection{Time-Constrained Information Gathering}\label{sec:ED}
To select a subset $ \mathcal{S} $ of models to observe,
we adopt an approach that builds on classical experiment design:
we suppose fitting each machine learning model $j \in [n]$ returns a linear measurement $x^T y_j$ of $x$, corrupted by Gaussian noise.
To estimate $x$, we would like to choose a set of observations $y_j$ that span $\mathbb{R}^k$
and form a well-conditioned submatrix, but that corresponds to models which are fast to run.
In passing, we note that the pivoted QR algorithm on the matrix $Y$
(heuristically) finds a well conditioned set of $k$ columns of $Y$.
However, we would like to find a method that is runtime-aware.

Our experiment design (ED) procedure minimizes
a scalarization of the covariance of the estimated meta-features $\hat{x}$ of the new dataset
subject to runtime constraints \cite{wald1943efficient,mood1946hotelling,john1975d,pukelsheim1993optimal,boyd2004convex}.
Formally, define an indicator vector $v \in \{0,1\}^n$,
where entry $v_j$ indicates whether to fit model $j$.
Let $\hat{t}_j$ denote the predicted runtime of model $j$ on a meta-test dataset,
and let $ y_j$ denote its latent meta-features, for $j \in [n]$.
Now relax to allow $v \in [0,1]^n$ to allow for non-Boolean values
and solve the optimization problem
\begin{equation}
\begin{array}{ll}
	\text{minimize} & \log \det \Big(\sum_{j=1}^n v_j y_j y_j^\top \Big)^{-1} \\
	\text{subject to} & \sum\limits_{j=1}^n v_j \hat{t}_j \leq \tau \\
                    & v_j \in [0, 1], \forall j \in [n]
\end{array}
\label{equation:ED_time}
\end{equation}
with variable $v \in \mathbb{R}^n$.
We call this method ED (time).
Scalarizing the covariance by minimizing the determinant is called D-optimal design.
Several other scalarizations can also be used, including covariance norm (E-optimal) or trace (A-optimal).
Replacing $t_i$ by 1 gives an alternative heuristic that bounds the \emph{number} of models fit by $\tau$;
we call this method ED (number).

Problem \ref{equation:ED_time} is a convex optimization problem,
and we obtain an approximate solution by rounding the largest entries of $v$ up to 1
until the selected models exceed the time limit $\tau$.
Let $\mathcal{S} \subseteq [n]$ be the set of indices of $e$ that we choose to observe, i.e. the set such that $v_s$ rounds to 1 for $ s \in \mathcal{S}$. We denote this process by $\mathcal{S} = \texttt{min\_variance\_ED}(\hat{t}, \{y_j\}_{j=1}^n, \tau)$.

\section{The \textsc{Oboe} system} \label{oboe}
\begin{figure*}
	\centering
	\includegraphics[width=.88\linewidth]{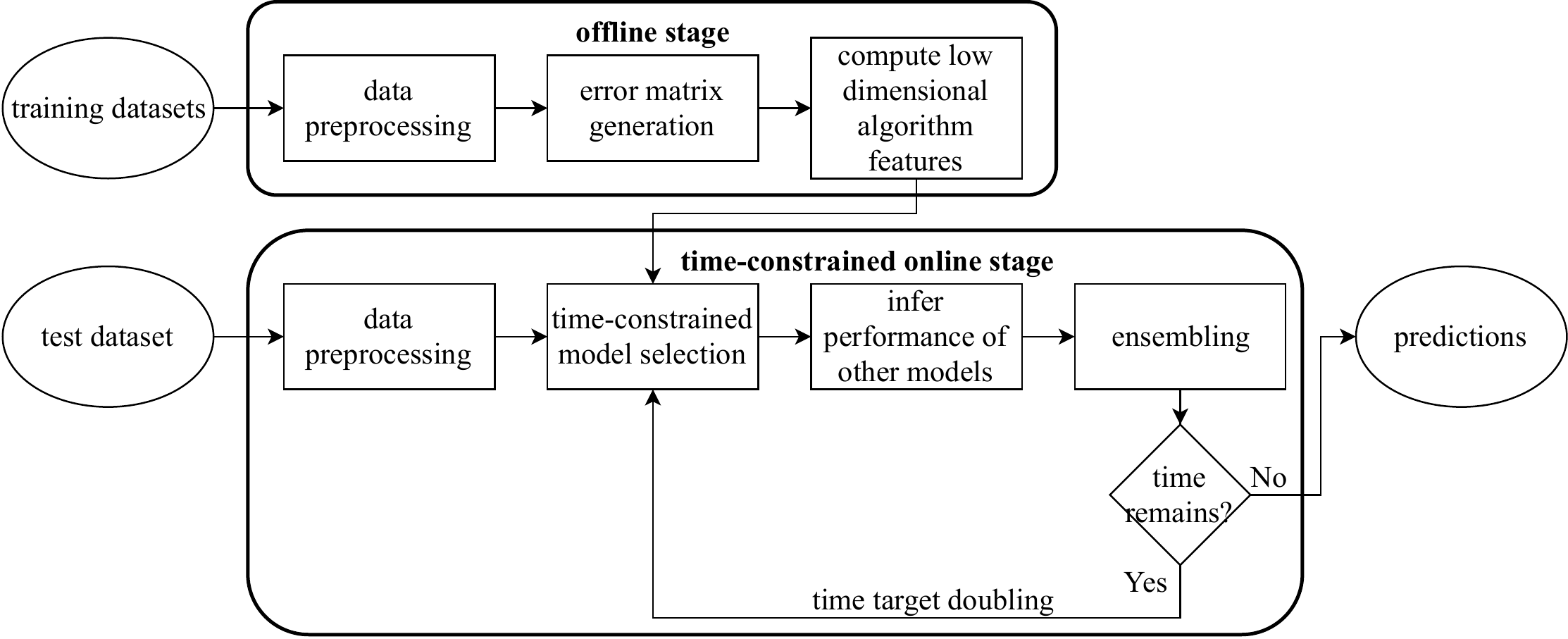}
	\caption{Diagram of data processing flow in the \textsc{Oboe} system.}
	\label{fig:oboe_diagram}
\end{figure*}

Shown in Figure~\ref{fig:oboe_diagram}, the \textsc{Oboe} system can be divided into offline and online stages.
The offline stage is executed only once and explores the space of model performance on meta-training datasets.
Time taken on this stage does not affect the runtime of \textsc{Oboe} on a new dataset;
the runtime experienced by user is that of the online stage.

One advantage of \textsc{Oboe} is that the vast majority of the time in the
online phase is spent training standard machine learning models,
while very little time is required to decide which models to sample.
Training these standard machine learning models requires running
algorithms on datasets with thousands of data points and features,
while the meta-learning task --- deciding which models to sample ---
requires only solving a small least-squares problem.

\subsection{Offline Stage}
The $(i,j)$th entry of error matrix $E \in \mathbb{R}^{m \times n}$, denoted as $E_{ij}$, records the performance of the $j$th model on the $i$th meta-training dataset.
We generate the error matrix using the \emph{balanced error rate} metric,
the average of false positive and false negative rates across different classes.
At the same time we record runtime of machine learning models on datasets. This is used to fit runtime predictors described in Section~\ref{methodology}.
Pseudocode for the offline stage is shown as Algorithm~\ref{alg:offline}.

\begin{algorithm}
	\caption{Offline Stage}
	\begin{algorithmic}[1]
		\Require{meta-training datasets $\{\mathcal{D}_i\}_{i=1}^m$, models $\{\mathcal{A}_j\}_{j=1}^n$, algorithm performance metric $\mathcal{M}$}

		\Ensure{error matrix $E$, runtime matrix $ T $, fitted runtime predictors $\{f_j\}_{j=1}^n$}

		\For{$i=1, 2, \ldots, m$}
		\State $ n^{\mathcal{D}_i}, p^{\mathcal{D}_i} \gets$ number of data points and features in $ \mathcal{D}_i $
		\For{$j=1,2,\ldots,n$}
		\State $E_{ij} \gets$ error of model $\mathcal{A}_j$ on dataset $\mathcal{D}_i$ according to metric $\mathcal{M}$
		\State $ T_{ij} \gets$ observed runtime for model $\mathcal{A}_j$ on dataset $\mathcal{D}_i$
		\EndFor
		\EndFor
		\For{$j=1, 2, \ldots, n$}
		\State fit $f_j = $ \texttt{fit\_runtime}$(n, p, T_j)$
		\EndFor
	\end{algorithmic}
	\label{alg:offline}
\end{algorithm}

\subsection{Online Stage} Recall that we repeatly double the time target of each round until we use up the total time budget. Thus each round is a subroutine of the entire online stage and is shown as Algorithm~\ref{alg:fit_one_round}, \texttt{fit\_one\_round}.

\begin{itemize}[leftmargin=*]
	\item \textbf{Time-constrained model selection (\texttt{fit\_one\_round})}
	Our active learning procedure selects a fast and informative collection of models to run on the meta-test dataset.
	\textsc{Oboe} uses the results of these fits to estimate the performance of all other models as accurately as possible.
	The procedure is as follows.
	First predict model runtime on the meta-test dataset using fitted runtime predictors.
	Then use experiment design to select a subset $\mathcal{S} $ of entries of $e$,
	the performance vector of the test dataset, to observe.
	The observed entries are used to compute $ \hat{x} $, an estimate of the latent meta-features of the test dataset,
	which in turn is used to predict every entry of $ e $.
	We build an ensemble out of models predicted to perform well within the time target $ \tilde{\tau} $
	by means of greedy forward selection \cite{caruana2004ensemble,caruana2006getting}.
	We denote this subroutine as $ \tilde{A}= $\texttt{ensemble\_selection}$(\mathcal{S}, e_\mathcal{S}, z_\mathcal{S})$,
	which takes as input the set of base learners $ \mathcal{S} $ with their cross-validation errors $ e_\mathcal{S} $ and predicted labels $ z_\mathcal{S} = \{z_s | s \in \mathcal{S}\}$,
	and outputs ensemble learner $ \tilde{A}$.
	The hyperparameters used by models in the ensemble can be tuned further, but in
	our experiments we did not observe substantial improvements from further hyperparameter tuning.

	\begin{algorithm}
		\caption{\texttt{fit\_one\_round}$(\{y_j\}_{j=1}^n, \{f_j\}_{j=1}^n, \mathcal{D}_{tr}, \tilde{\tau})$}
		\begin{algorithmic}[1]

			\Require{model latent meta-features $ \{y_j\}_{j=1}^n $, fitted runtime predictors $\{f_j\}_{j=1}^n$, training fold of the meta-test dataset $\mathcal{D}_\text{tr}$, number of best models $ N $ to select from the estimated performance vector, time target for this round $\tilde{\tau}$}
			\Ensure{ensemble learner $ \tilde{A}$}

			\For{$j=1, 2, \ldots, n$}
			\State $\hat{t}_j \gets f_j(n^{\mathcal{D}_\text{tr}}, p^{\mathcal{D}_\text{tr}})$
			\EndFor
			\State $\mathcal{S} = \texttt{min\_variance\_ED}(\hat{t}, \{y_j\}_{j=1}^n, \tilde{\tau})$
			\For{$k=1, 2, \ldots, |\mathcal{S}|$}
			\State $e_{\mathcal{S}_k} \gets$ cross-validation error of model $ \mathcal{A}_{\mathcal{S}_k} $ on $ \mathcal{D}_\text{tr} $
			\EndFor
			\State $ \hat{x} \gets (\begin{bmatrix} y_{\mathcal{S}_1} & y_{\mathcal{S}_2} & \cdots & y_{\mathcal{S}_{|\mathcal{S}|}} \end{bmatrix}^\top)^\dagger e_S$
			\State $ \hat{e} \gets \begin{bmatrix} y_1 & y_2 & \cdots & y_n \end{bmatrix}^\top \hat{x}$
			\State $ \mathcal{T} \gets $ the $ N $ models with lowest predicted errors in $ \hat{e} $
			\For{$k=1, 2, \ldots, |\mathcal{T}|$}
			\State $e_{\mathcal{T}_k}, z_{\mathcal{T}_k} \gets$ cross-validation error of model $ \mathcal{A}_{\mathcal{T}_k} $ on $ \mathcal{D}_\text{tr} $
			\EndFor
			\State $\tilde{A} \gets $\texttt{ensemble\_selection}$(\mathcal{T}, e_\mathcal{T}, z_\mathcal{T})$
		\end{algorithmic}
		\label{alg:fit_one_round}
	\end{algorithm}

	\item \textbf{Time target doubling}\label{s-doubling}
	To select rank $ k $, \textsc{Oboe} starts with a small initial rank along with a small time target,
	and then doubles the time target for \texttt{fit\_one\_round} until the elapsed time reaches half of the total budget.
	The rank $ k $ increments by 1 if the validation error of the ensemble learner
	decreases after doubling the time target, and otherwise does not change.
	Since the matrices returned by PCA with rank $k$ are submatrices of those returned by PCA with rank $l$ for $l > k$, we can compute the factors as submatrices of the $ \textit{m} $-by-$ \textit{n} $ matrices returned by PCA with full rank $\min(m,n)$ \cite{golub2012matrix}.
	The pseudocode is shown as Algorithm~\ref{alg:onlinefitdoubling}.
\end{itemize}

\begin{algorithm}
	\caption{Online Stage}
	\begin{algorithmic}[1]
		\Require{error matrix $E$, runtime matrix $T$, meta-test dataset $\mathcal{D}$, total time budget $\tau$, fitted runtime predictors $\{f_j\}_{j=1}^n$}, initial time target $\tilde{\tau}_0$, initial approximate rank $ k_0 $

		\Ensure{ensemble learner $ \tilde{A} $}

		\State $x_i, y_j \gets \argmin \sum_{i=1}^m \sum_{j=1}^n (E_{ij} - x_i^\top y_j)^2$, $x_i \in \mathbb{R}^{\min(m,n)}$ for $i \in [M]$ , $y_j \in \mathbb{R}^{\min(m,n)}$ for $j \in [N]$
		\State $ \mathcal{D}_\text{tr}, \mathcal{D}_\text{val}, \mathcal{D}_\text{te} \gets$ training, validation and test folds of $ \mathcal{D} $
		\State $ \tilde{\tau} \gets \tilde{\tau}_0$
		\State $ k \gets k_0$
		\While{$\tilde{\tau} \leq \tau/2$}
		\State $\{\tilde{y}_j\}_{j=1}^n \gets$ \textit{k}-dimensional subvectors of $ \{y_j\}_{j=1}^n $
		\State $ \tilde{A} \gets$ \texttt{fit\_one\_round}$(\{\tilde{y}_j\}_{j=1}^n, \{f_j\}_{j=1}^n, \mathcal{D}_\text{tr}, \tilde{\tau})$
		\State $ e_{\tilde{A}}' \gets \tilde{A}(\mathcal{D}_\text{val})$
		\If {$ e_{\tilde{A}}' < e_{\tilde{A}} $ }
		\State $ k \gets k+1 $
		\EndIf
		\State $ \tilde{\tau} \gets 2\tilde{\tau}$
		\State $ e_{\tilde{A}} \gets e_{\tilde{A}}' $

		\EndWhile
	\end{algorithmic}
	\label{alg:onlinefitdoubling}
\end{algorithm}

\section{Experimental evaluation} \label{experiments}
We ran all experiments on a server with 128 Intel\textsuperscript{\textregistered} Xeon\textsuperscript{\textregistered} E7-4850 v4 2.10GHz CPU cores.
The process of running each system on a specific dataset is limited to a single CPU core.
Code for the \textsc{Oboe} system is at \url{https://github.com/udellgroup/oboe}; code for experiments is at \url{https://github.com/udellgroup/oboe-testing}.

We test different AutoML systems on midsize OpenML and UCI datasets, using standard machine learning models shown in Appendix~\ref{supp:models}, Table~\ref{table:models}.
Since data pre-processing is not our focus, we pre-process all datasets in the same way: one-hot encode categorical features and then standardize all features to have zero mean and unit variance.
These pre-processed datasets are used in all the experiments.

\subsection{Performance Comparison across AutoML Systems}
We compare AutoML systems that are able to select among different algorithm types under time constraints: \textsc{Oboe} (with error matrix generated from midsize OpenML datasets), auto-sklearn \cite{feurer2015efficient}, probabilistic matrix factorization (PMF) \cite{fusi2018probabilistic},
and a \textit{time-constrained} random baseline.
The time-constrained random baseline selects models to observe randomly
from those predicted to take less time than the remaining time budget
until the time limit is reached.

\subsubsection{Comparison with PMF}
\label{section:numerics_of_comparison_with_pmf}
PMF and \textsc{Oboe} differ in the surrogate models they use to explore the model space:
PMF incrementally picks models to observe using Bayesian optimization, with model latent meta-features from probabilistic matrix factorization as features,
while \textsc{Oboe} models algorithm performance as bilinear in model and dataset meta-features.

\begin{figure}
	\centering
	\begin{subfigure}[t]{.47\linewidth}
		\includegraphics[width=\linewidth]{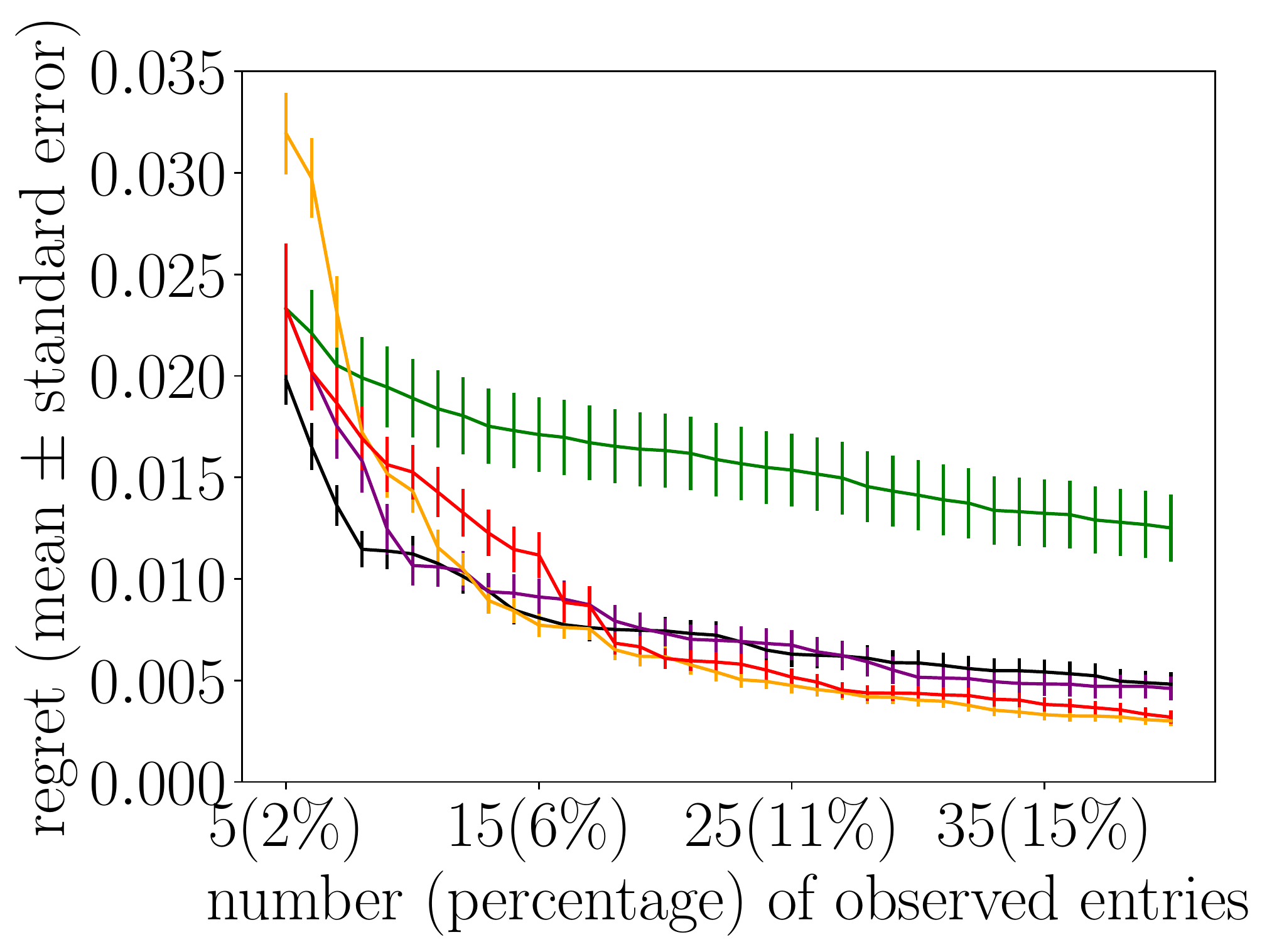}
	\end{subfigure}%
\begin{subfigure}[t]{.47\linewidth}
	\includegraphics[width=\linewidth]{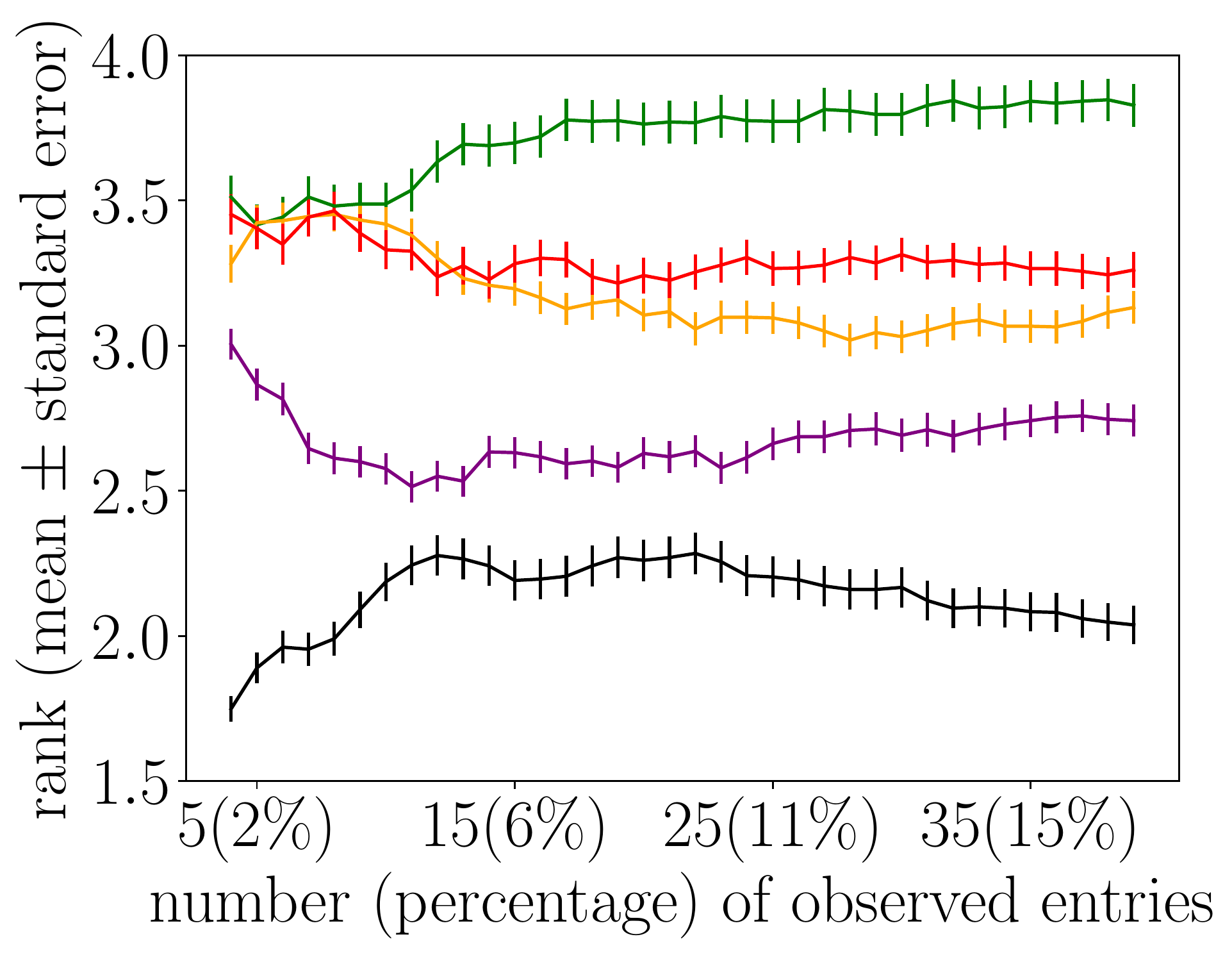}
\end{subfigure}%

\begin{subfigure}[t]{.9\linewidth}
		\includegraphics[width=\linewidth]{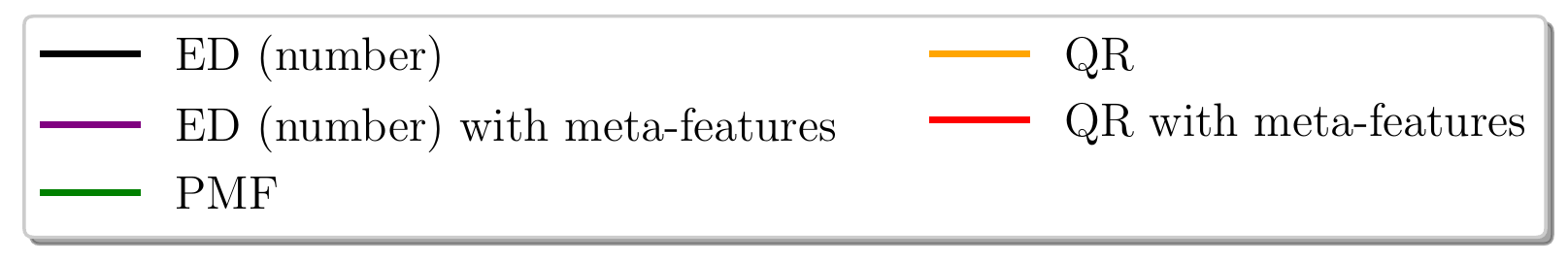}
\end{subfigure}

	\caption{Comparison of sampling schemes (QR or ED) in \textsc{Oboe} and PMF.
	"QR" denotes QR decomposition with column pivoting;
	"ED (number)" denotes experiment design with number of observed entries constrained.
	The left plot shows the regret of each AutoML method as a function of number of entries;
	the right shows the relative rank of each AutoML method in the regret plot (1 is best and 5 is worst).
	}
	\label{fig:comparison_with_pmf}
\end{figure}

PMF does not limit runtime, hence we compare it to \textsc{Oboe} using
either QR or ED (number) to decide the set $S$ of models (see Section~\ref{sec:ED}).
Figure~\ref{fig:comparison_with_pmf} compares the performance of PMF and
\textsc{Oboe} (using QR and ED (number) to decide the set $S$ of models)
on our collected error matrix
to see which is best able to predict the smallest entry in each row.
 We show the regret: the difference between the minimal entry in each row and the one found by the AutoML method.
In PMF, $N_0 =5$ models are chosen from the best algorithms on similar datasets
(according to dataset meta-features shown in Appendix~\ref{supp:metafeature}, Table~\ref{table:metafeatures})
are used to warm-start Bayesian optimization, which then searches for the next model to observe.
\textsc{Oboe} does not require this initial information before beginning its exploration.
However, for a fair comparison, we show both "warm" and "cold" versions.
The warm version observes both the models chosen by meta-features and those chosen by QR or ED;
the number of observed entries in Figure~\ref{fig:comparison_with_pmf} is the sum of all observed models.
The cold version starts from scratch and only observes models chosen by QR and ED.

(Standard ED also performs well; see Appendix~\ref{sec:comparison_of_different_ED}, Figure~\ref{fig:comparison_of_different_ED}.)

Figure~\ref{fig:comparison_with_pmf} shows the surprising effectiveness of
the low rank model used by \textsc{Oboe}:
\begin{enumerate}[label=\arabic*, wide, labelwidth=!, labelindent=0pt]

	\item Meta-features are of marginal value in choosing new models to observe.
	For QR, using models chosen by meta-features helps when the number of observed entries is small.
	For ED, there is no benefit to using models chosen by meta-features.

	\item The low rank structure used by QR and ED seems to provide a better guide
	to which models will be informative than the Gaussian process prior used by PMF:
	the regret of PMF does not decrease as fast as \textsc{Oboe} using either QR or ED.
\end{enumerate}

\subsubsection{Comparison with auto-sklearn}
\label{comparison_with_auto_sklearn}
The comparison with PMF assumes we can use the labels for every point in the entire dataset for model selection, so we can compare the performance of every model selected
and pick the one with lowest error.
In contrast, our comparison with auto-sklearn takes place in a more challenging, realistic setting: when doing cross-validation on the meta-test dataset, we do not know the labels of the validation fold until we evaluate performance of the ensemble we built within time constraints on the training fold.

 Figure~\ref{fig:comparison_with_autosklearn} shows the error rate and ranking of
 each AutoML method as the runtime repeatedly doubles.
 Again, \textsc{Oboe}'s simple bilinear model performs surprisingly well'\footnote{Auto-sklearn's GitHub Issue \#537 says ``Do not start auto-sklearn for time limits less than 60s". These plots should not be taken as criticisms of auto-sklearn,
 but are used to demonstrate \textsc{Oboe}'s ability to select a model within a short time.}:

\begin{figure}
	\centering
	\begin{subfigure}[t]{.47\linewidth}
		\includegraphics[width=\linewidth]{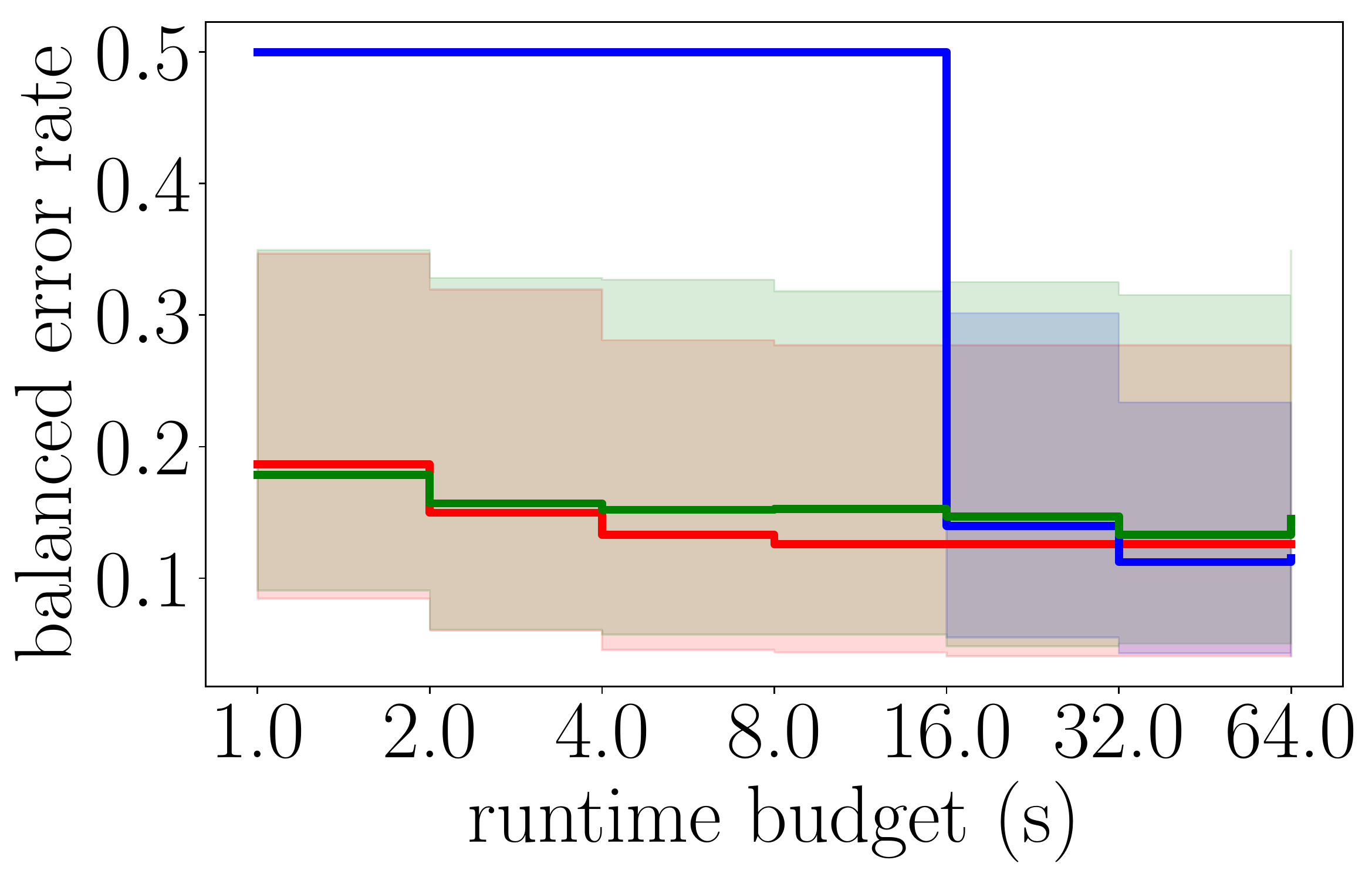}
		\caption{OpenML (meta-LOOCV)}
		\label{percentile_OpenML}
	\end{subfigure}%
	\begin{subfigure}[t]{.47\linewidth}
		\includegraphics[width=\linewidth]{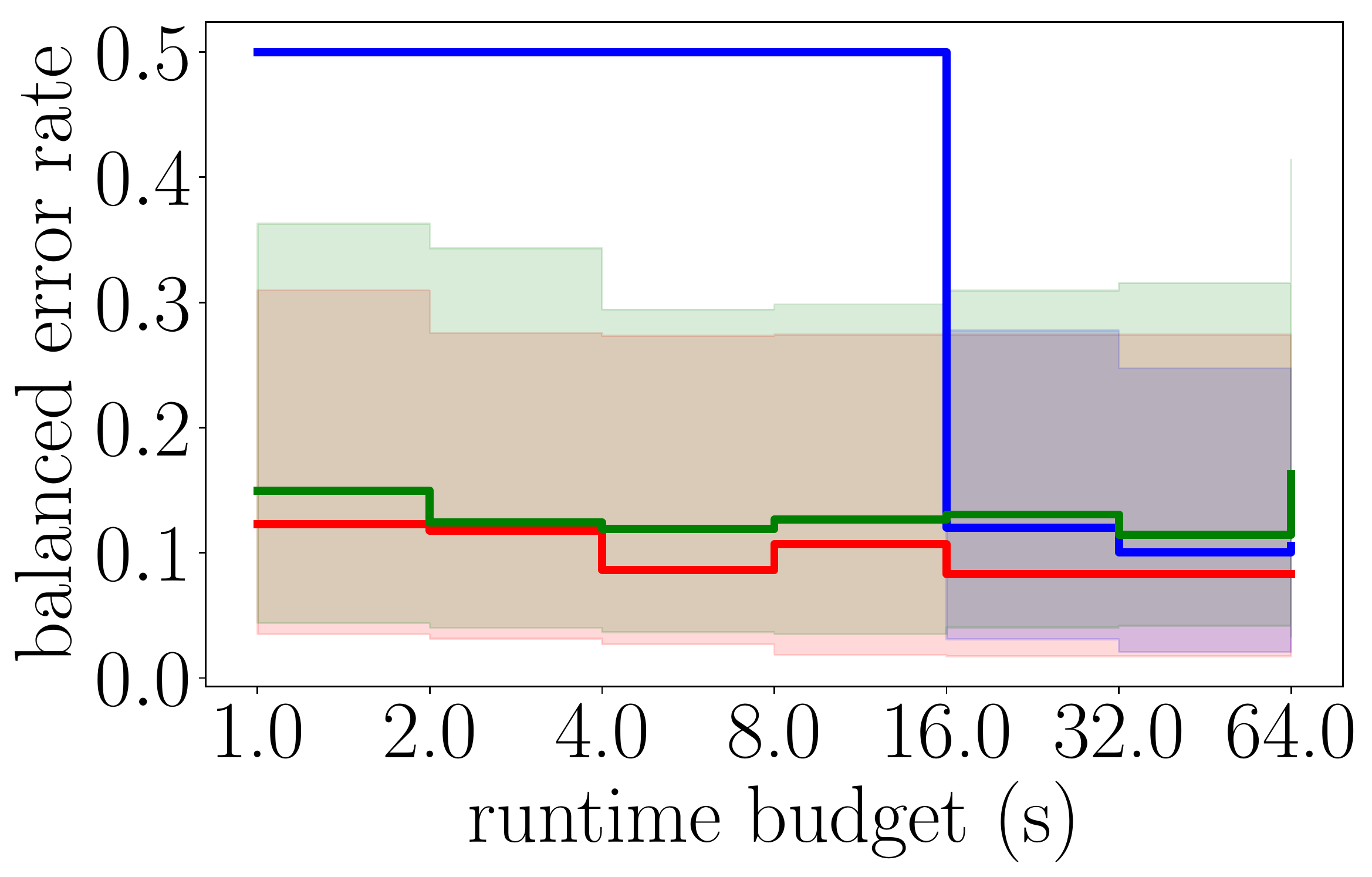}
		\caption{UCI (meta-test)}
		\label{percentile_UCI}
	\end{subfigure}%

	\begin{subfigure}[t]{0.47\linewidth}
		\includegraphics[width=\linewidth]{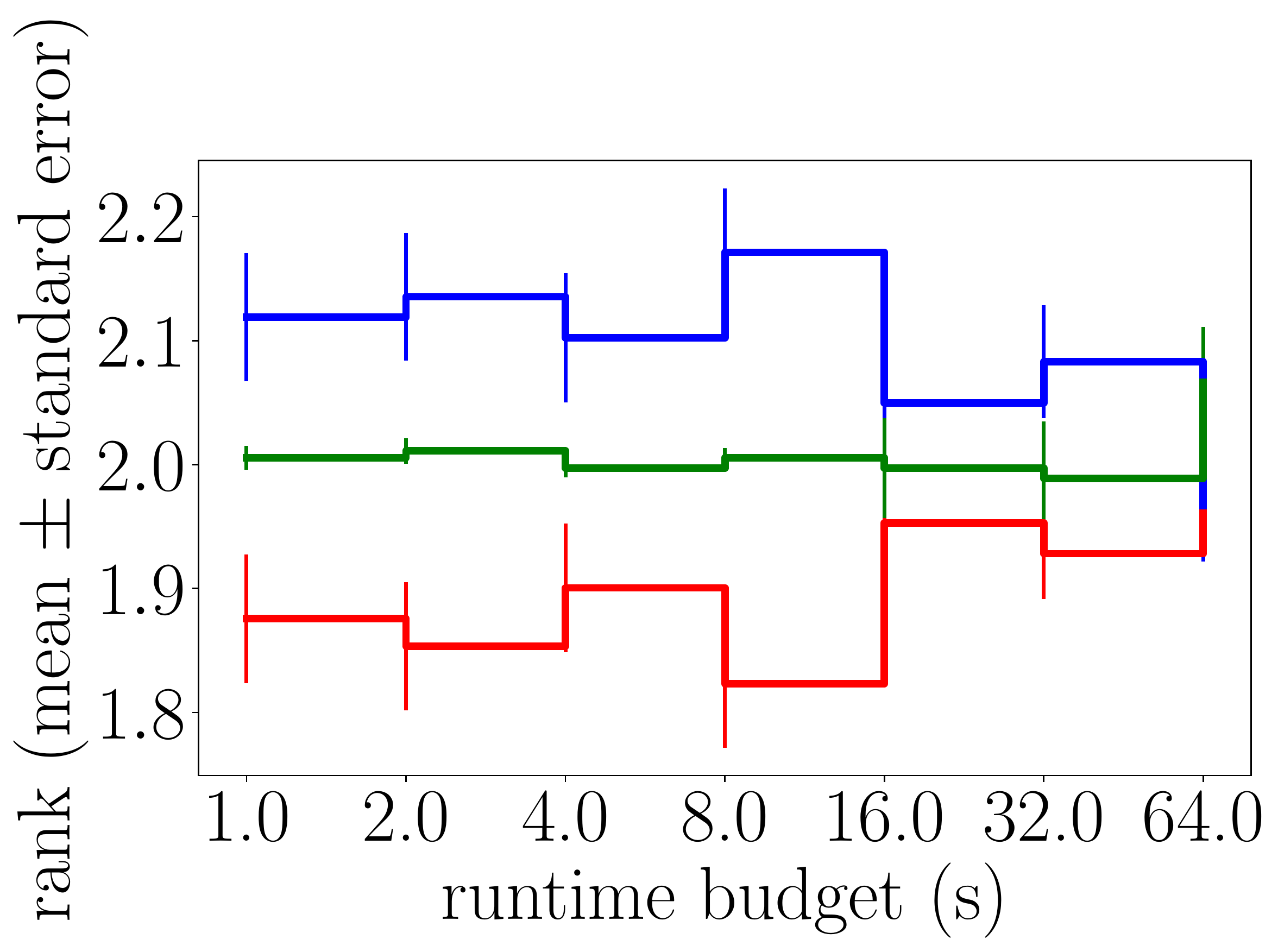}
		\caption{\label{rank-openml} OpenML (meta-LOOCV)}
	\end{subfigure}
	\begin{subfigure}[t]{0.47\linewidth}
		\includegraphics[width=\linewidth]{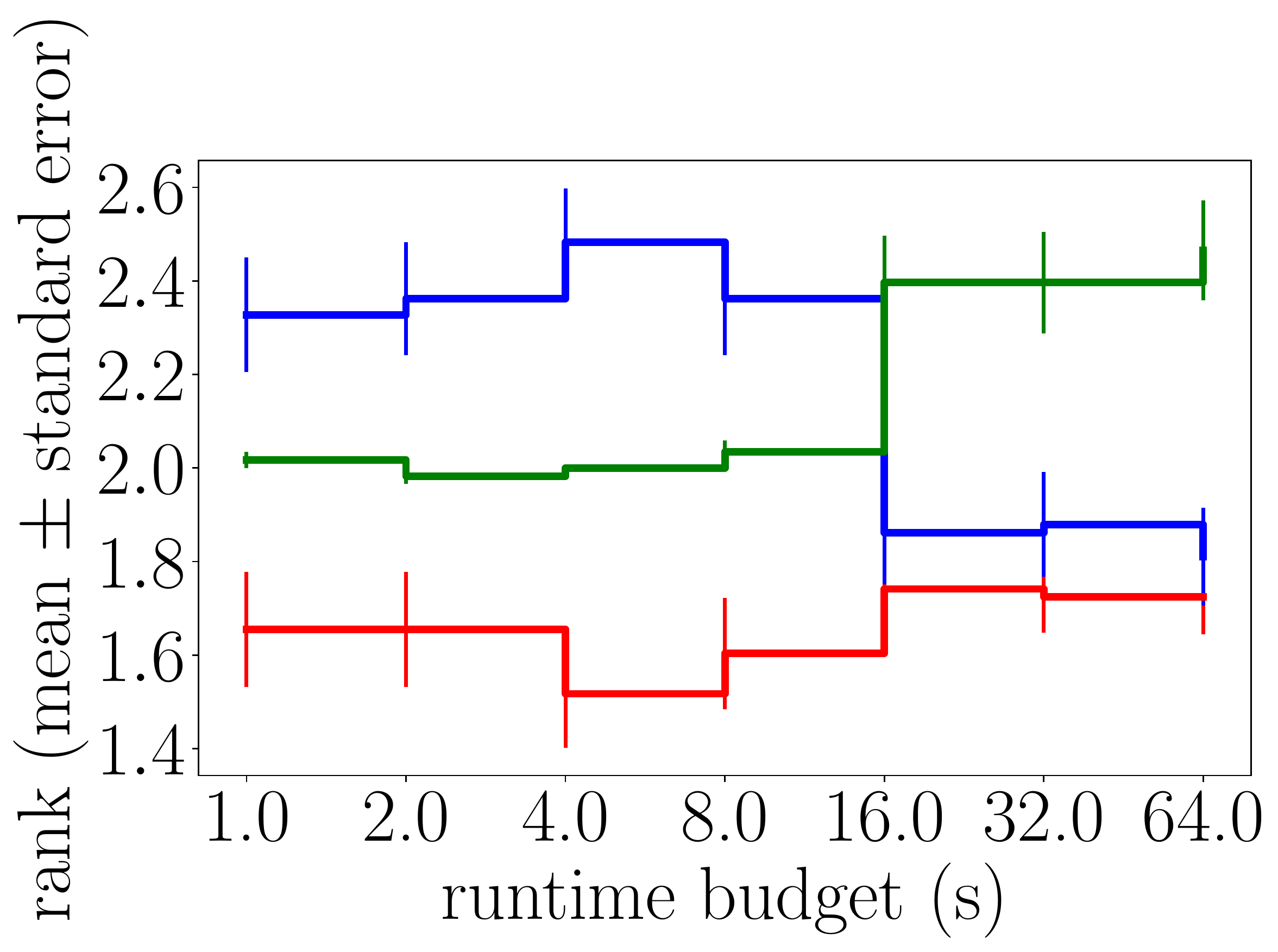}
		\caption{\label{rank-UCI} UCI (meta-test)}
	\end{subfigure}
	\caption{
		Comparison of AutoML systems in a time-constrained setting, including \textsc{Oboe} with experiment design (red), auto-sklearn (blue), and \textsc{Oboe} with time-constrained random initializations (green).
		OpenML and UCI denote midsize OpenML and UCI datasets.
		"meta-LOOCV" denotes leave-one-out cross-validation across datasets.
		In~\ref{percentile_OpenML} and~\ref{percentile_UCI}, solid lines represent medians;
		shaded areas with corresponding colors represent the regions between 75th and 25th percentiles.
		Until the first time the system can produce a model,
		we classify every data point with the most common class label.
		Figures~\ref{rank-openml} and~\ref{rank-UCI}
		show system rankings (1 is best and 3 is worst).}
	\label{fig:comparison_with_autosklearn}
\end{figure}

\begin{enumerate}[label=\arabic*, wide, labelwidth=!, labelindent=0pt]
	\item \textsc{Oboe} on average performs as
	well as or better than auto-sklearn (Figures~\ref{rank-openml} and~\ref{rank-UCI}).

	\item The quality of the initial models computed by \textsc{Oboe} and by auto-sklearn are comparable,
	but \textsc{Oboe} computes its first nontrivial model more than 8$ \times $ faster than auto-sklearn (Figures~\ref{percentile_OpenML} and~\ref{percentile_UCI}).
	In contrast, auto-sklearn must first compute meta-features for each dataset,
	which requires substantial computational time, as shown in Appendix~\ref{metafeaturetime}, Figure~\ref{fig:metafeature_calculation}.

	\item
	Interestingly, the rate at which the \textsc{Oboe} models improves with time is
	also faster than that of auto-sklearn: the improvement \textsc{Oboe} makes before 16s
	matches that of auto-sklearn from 16s to 64s.
	This indicates that the large time budget may be better spent in fitting more models
	than optimizing over hyperparameters, to which auto-sklearn devotes the remaining time.

	\item Experiment design leads to better results than random selection in almost all cases.

\end{enumerate}

\subsection{Why does \textsc{Oboe} Work?}

\begin{figure*}
		\begin{subfigure}[b]{0.33\linewidth}
			\includegraphics[width=\linewidth]{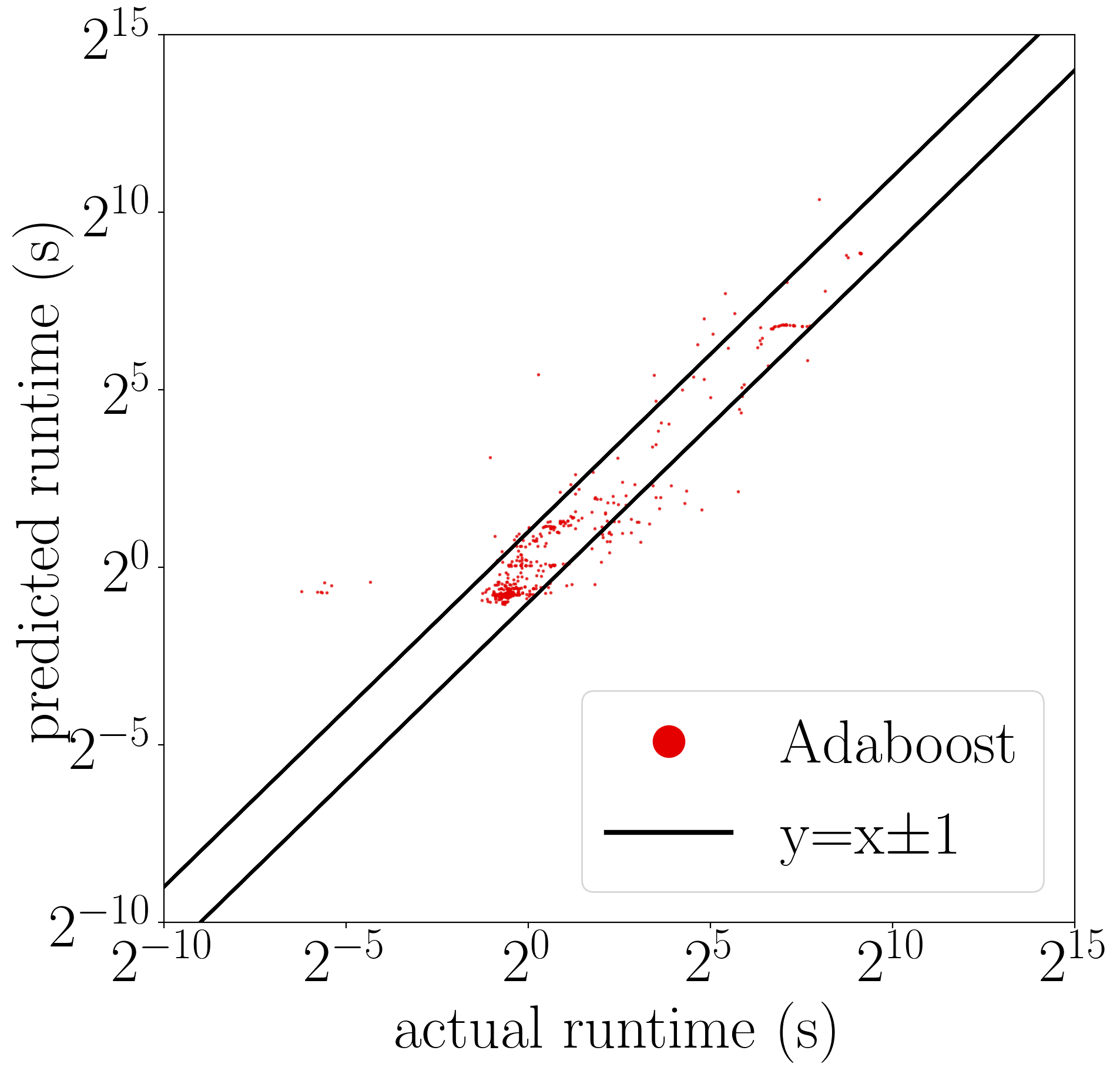}
		\end{subfigure}%
		\begin{subfigure}[b]{0.33\linewidth}
			\includegraphics[width=\linewidth]{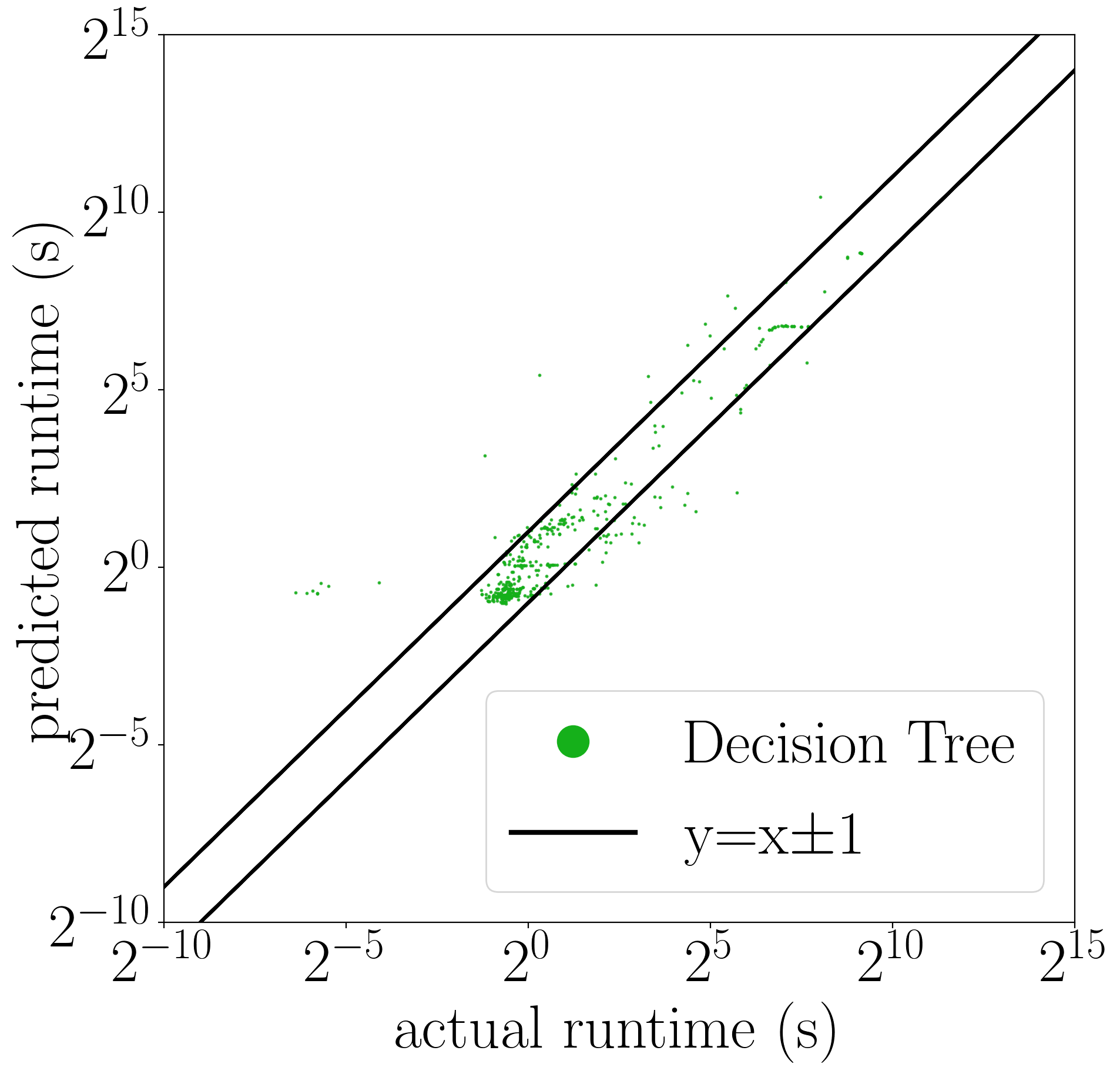}
		\end{subfigure}%
		\begin{subfigure}[b]{0.33\linewidth}
			\includegraphics[width=\linewidth]{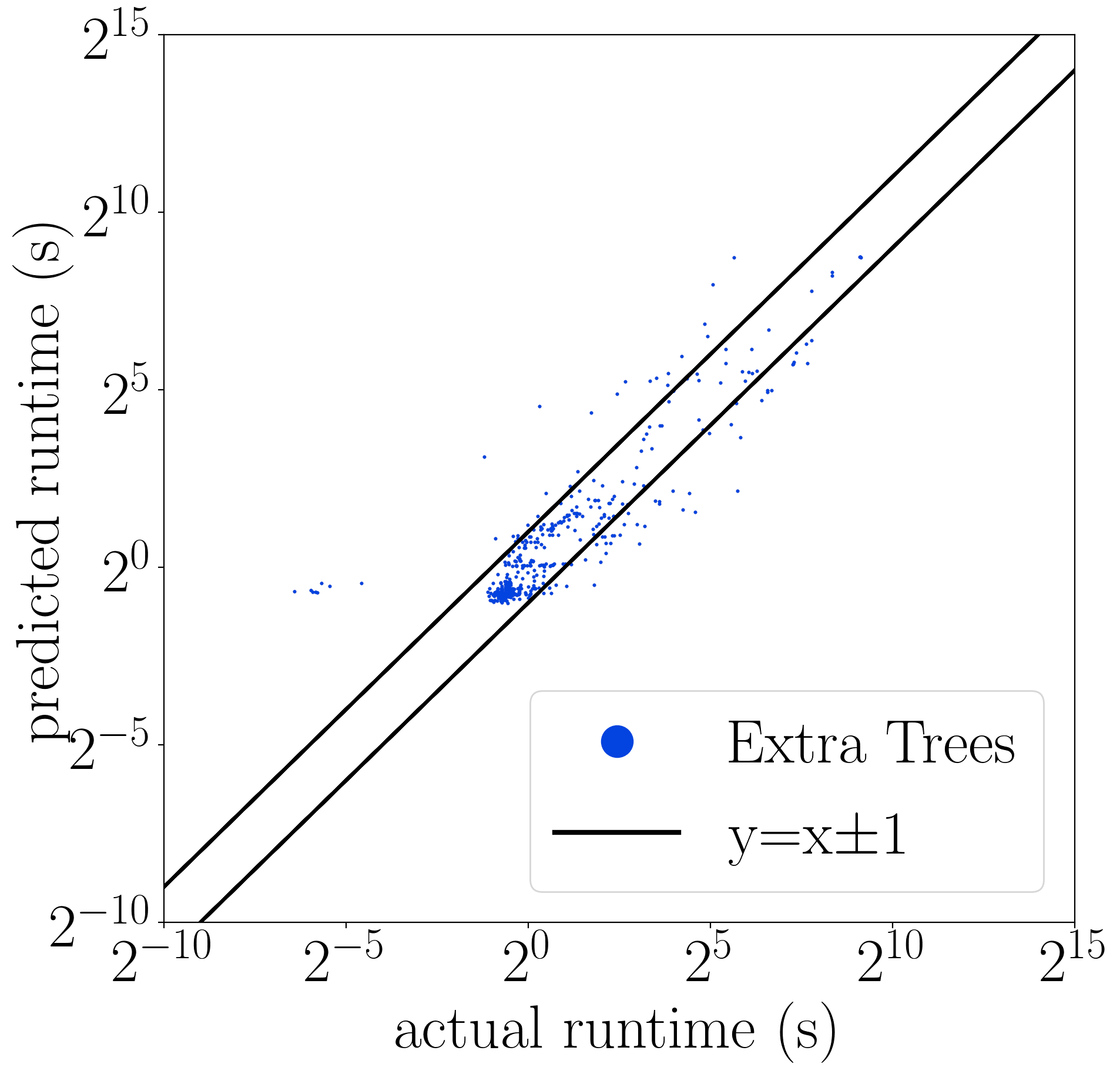}
		\end{subfigure}%

		\begin{subfigure}[b]{0.33\linewidth}
			\includegraphics[width=\linewidth]{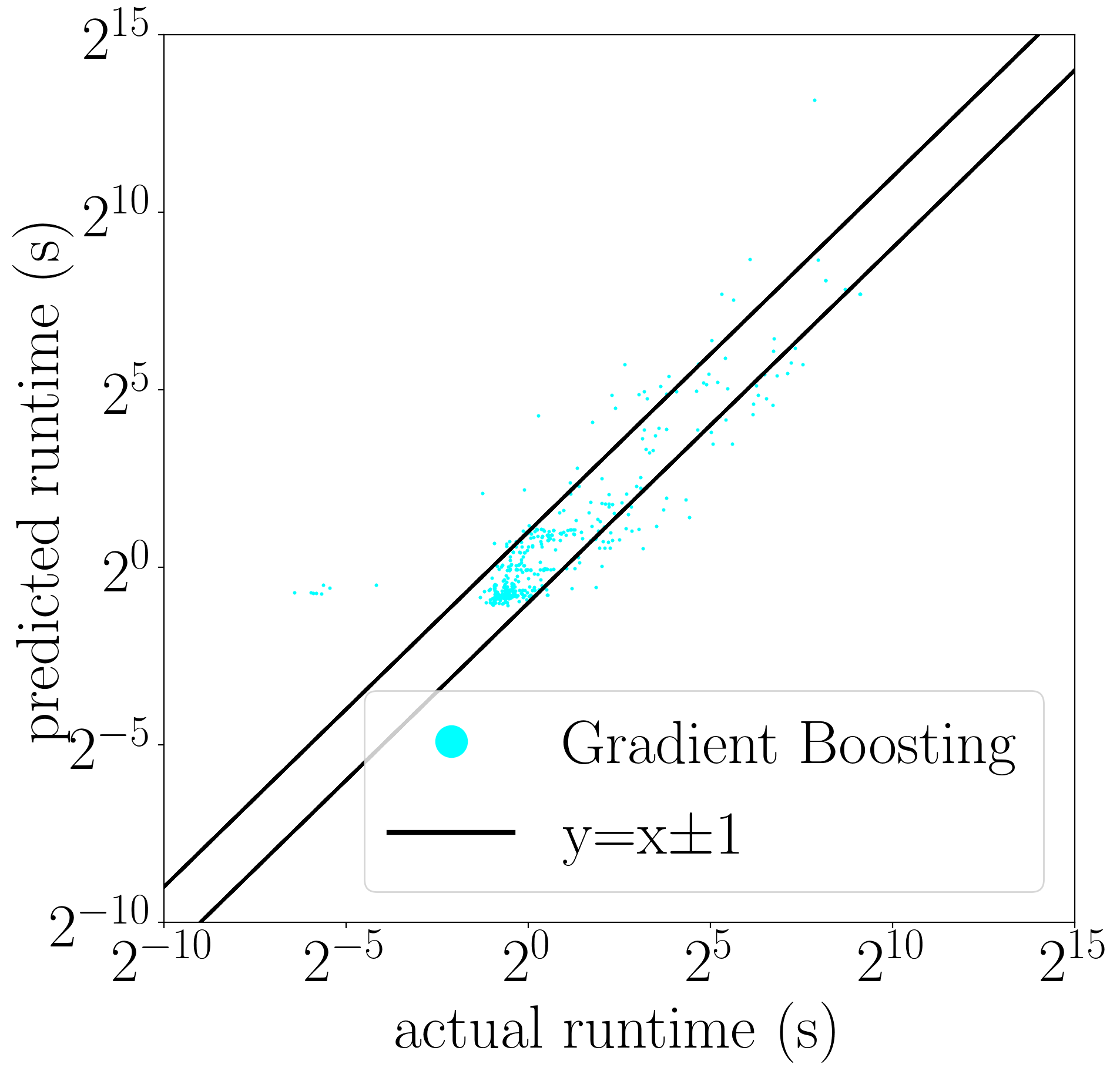}
		\end{subfigure}%
		\begin{subfigure}[b]{0.33\linewidth}
			\includegraphics[width=\linewidth]{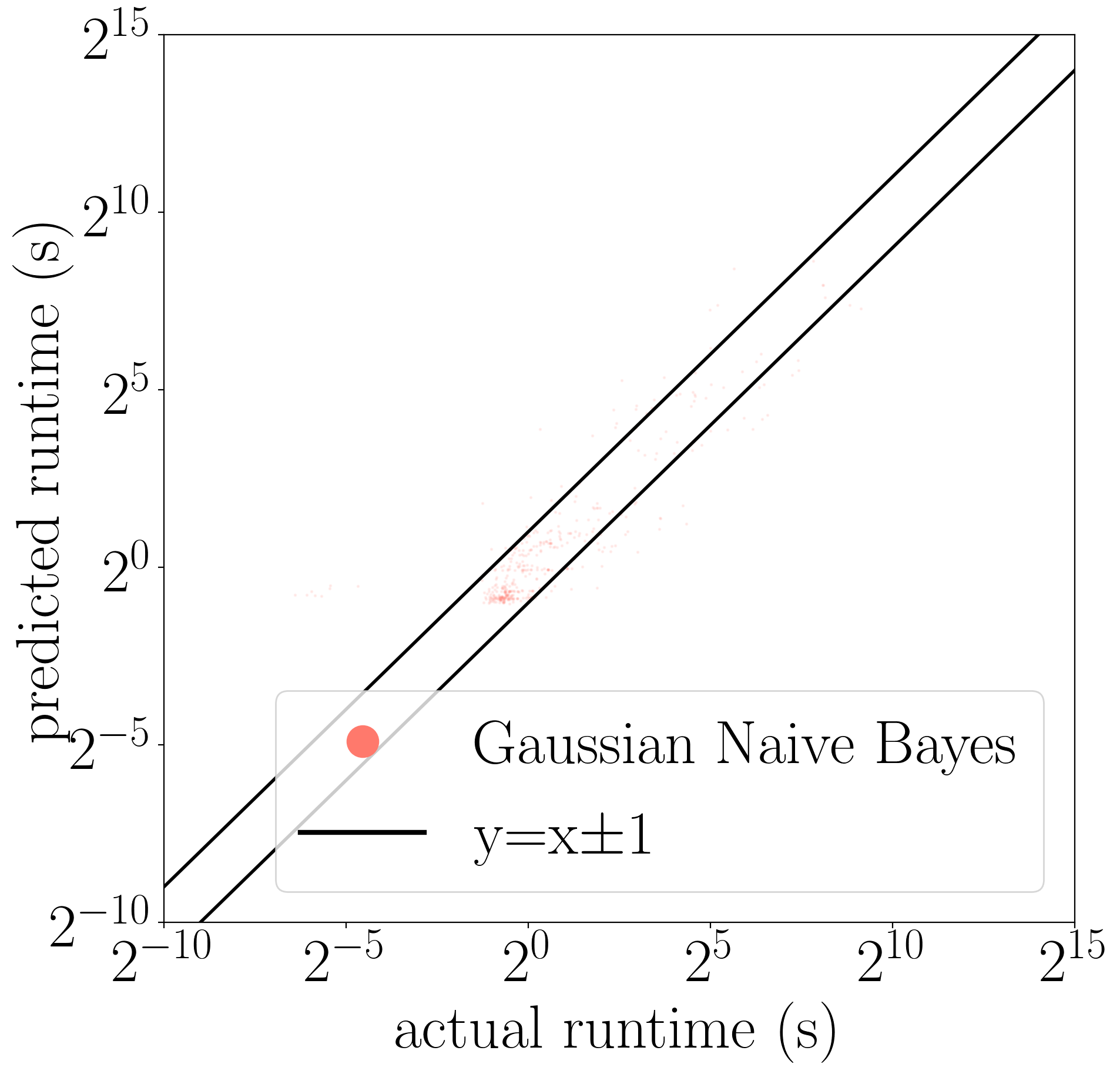}
		\end{subfigure}%
		\begin{subfigure}[b]{0.33\linewidth}
			\includegraphics[width=\linewidth]{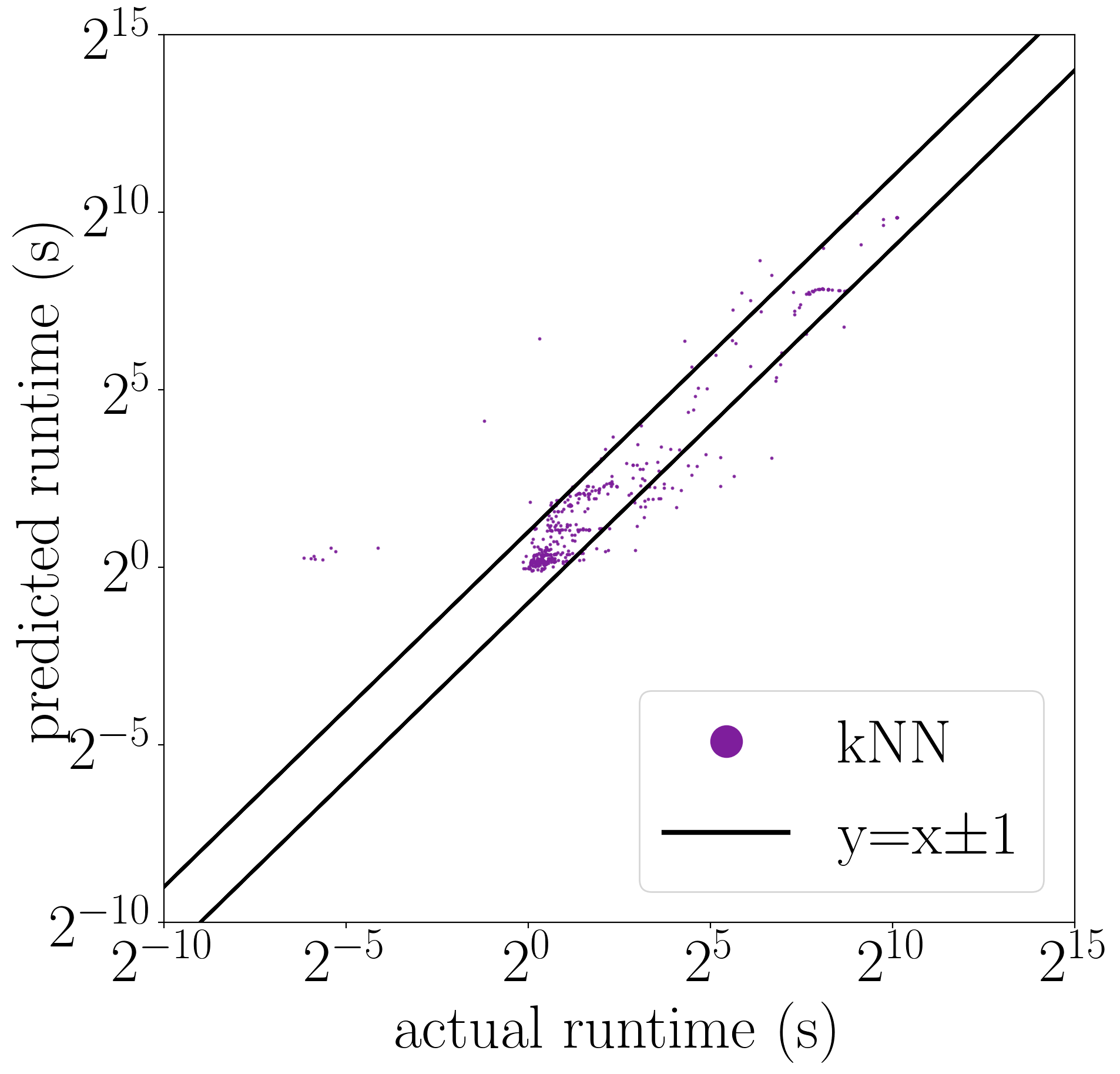}
		\end{subfigure}%

		\begin{subfigure}[b]{0.33\linewidth}
			\includegraphics[width=\linewidth]{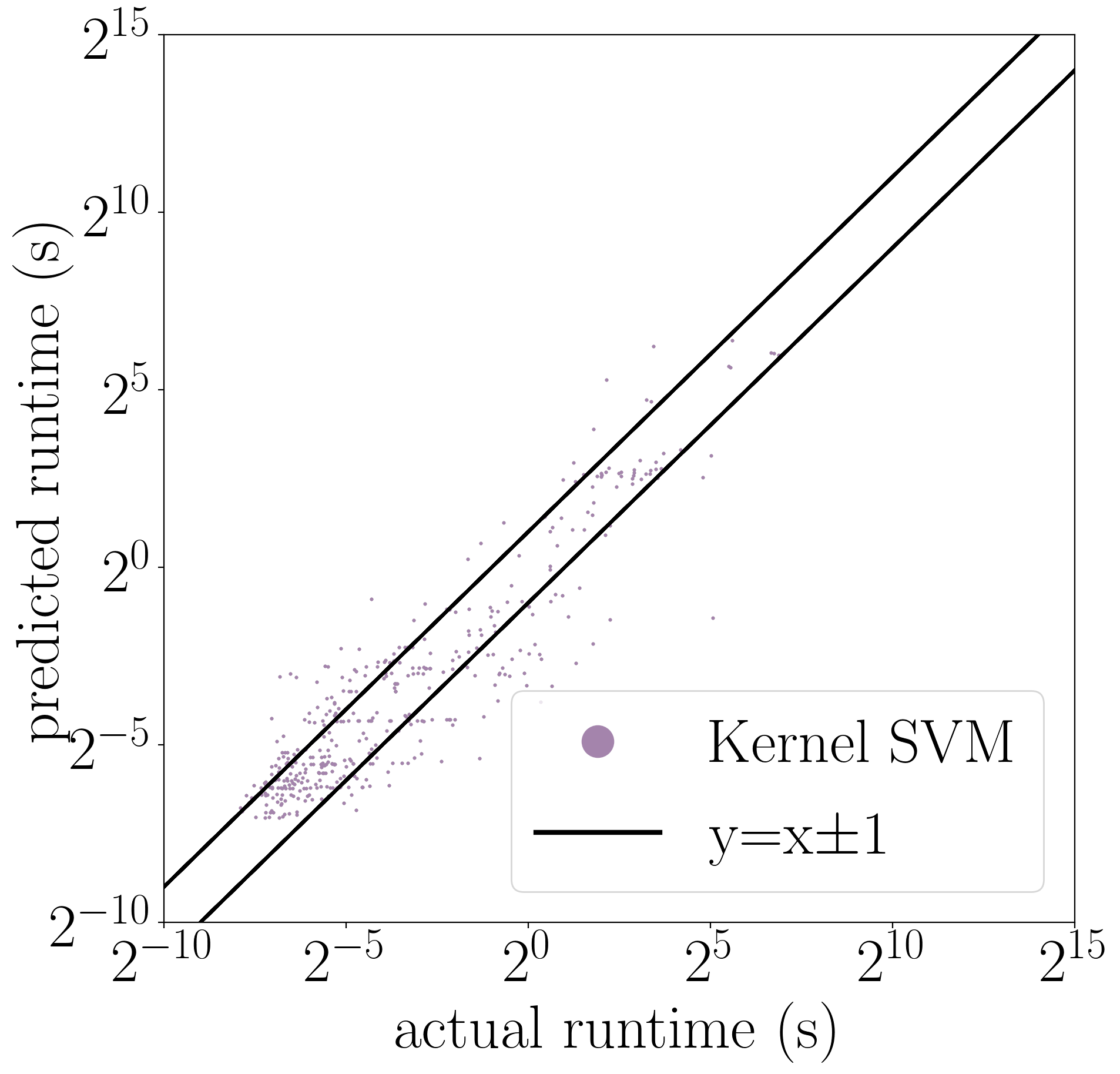}
		\end{subfigure}%
		\begin{subfigure}[b]{0.33\linewidth}
			\includegraphics[width=\linewidth]{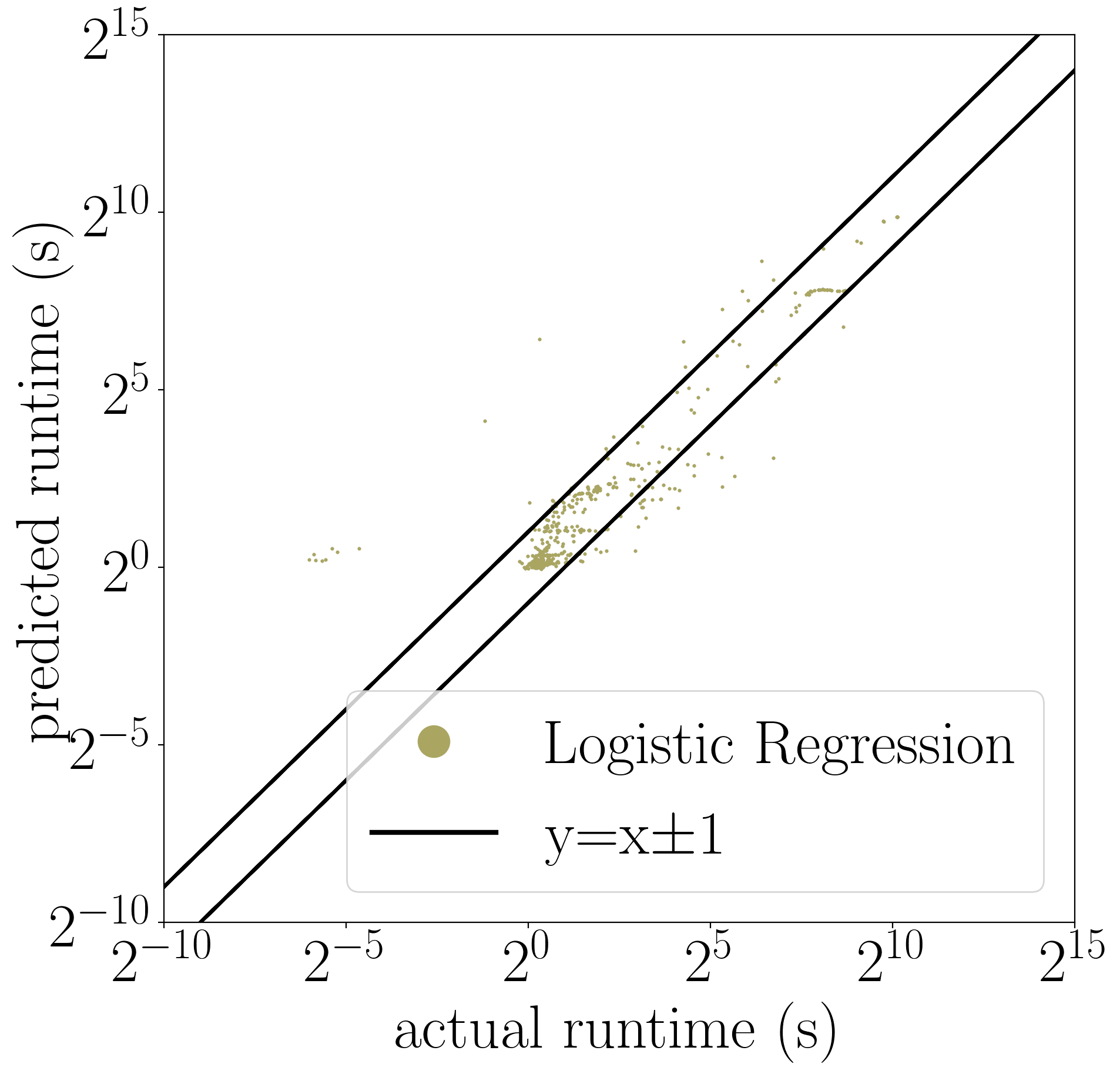}
		\end{subfigure}%
		\begin{subfigure}[b]{0.33\linewidth}
			\includegraphics[width=\linewidth]{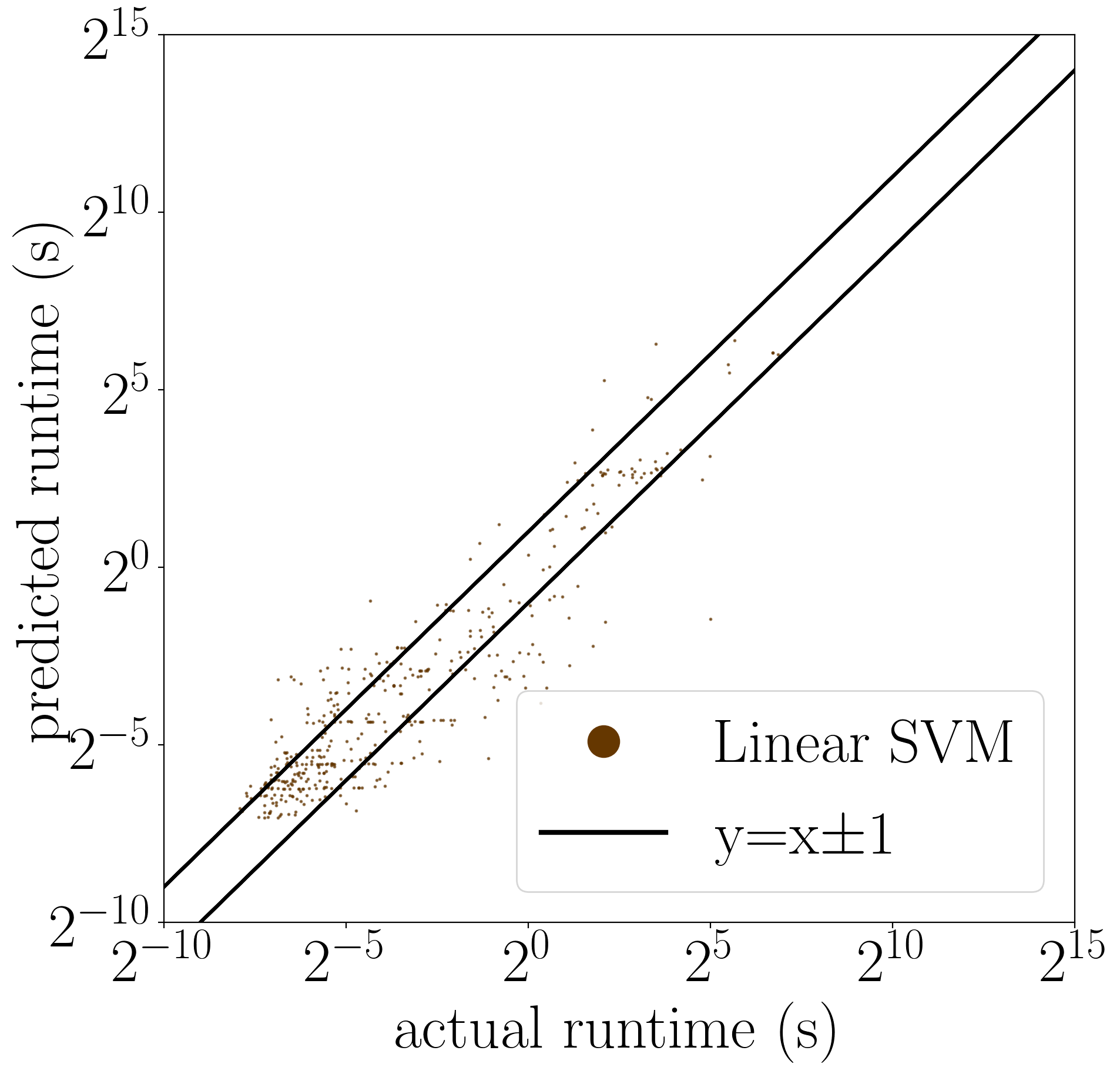}
		\end{subfigure}%

		\begin{subfigure}[b]{0.33\linewidth}
			\includegraphics[width=\linewidth]{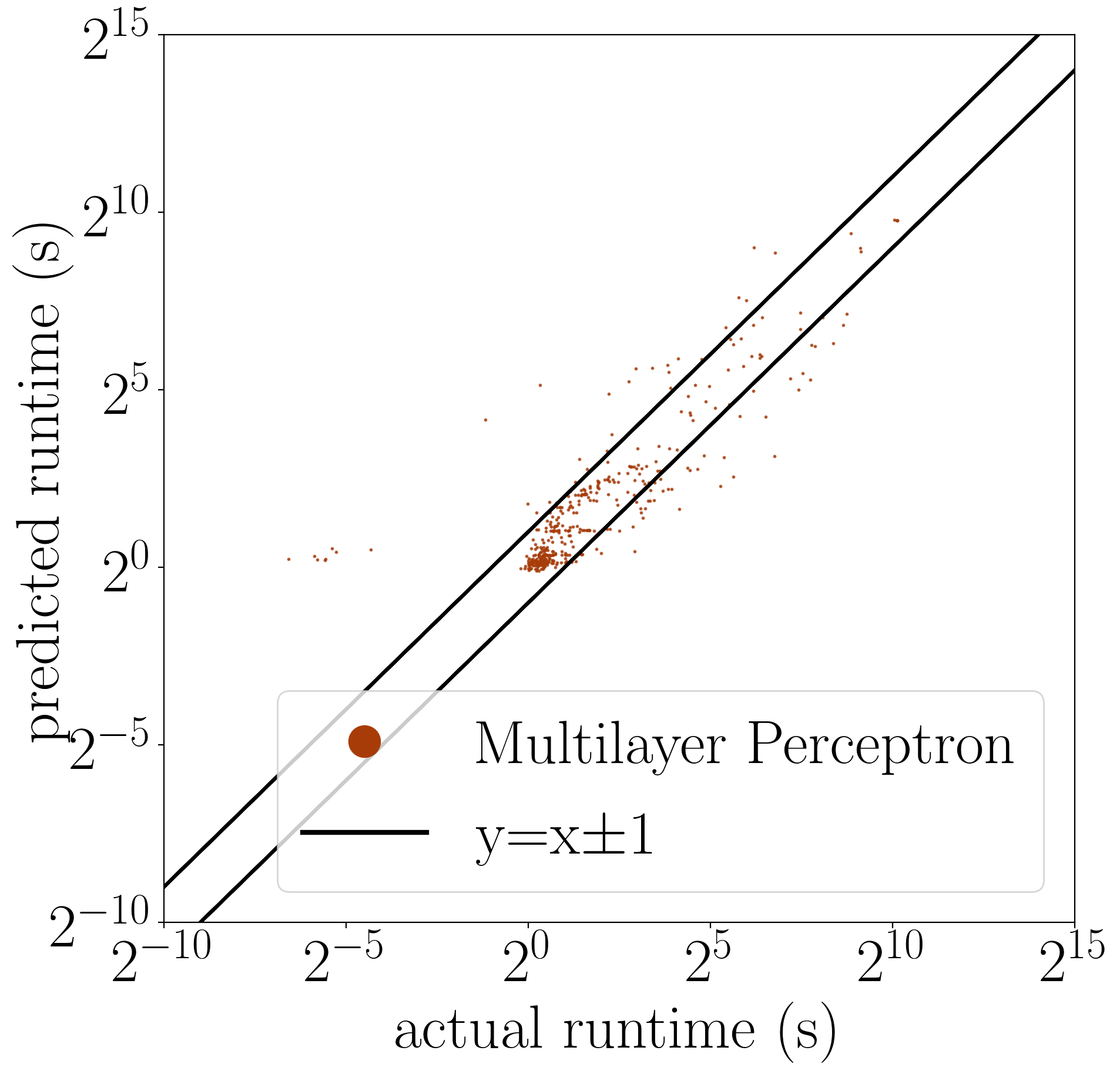}
		\end{subfigure}%
		\begin{subfigure}[b]{0.33\linewidth}
			\includegraphics[width=\linewidth]{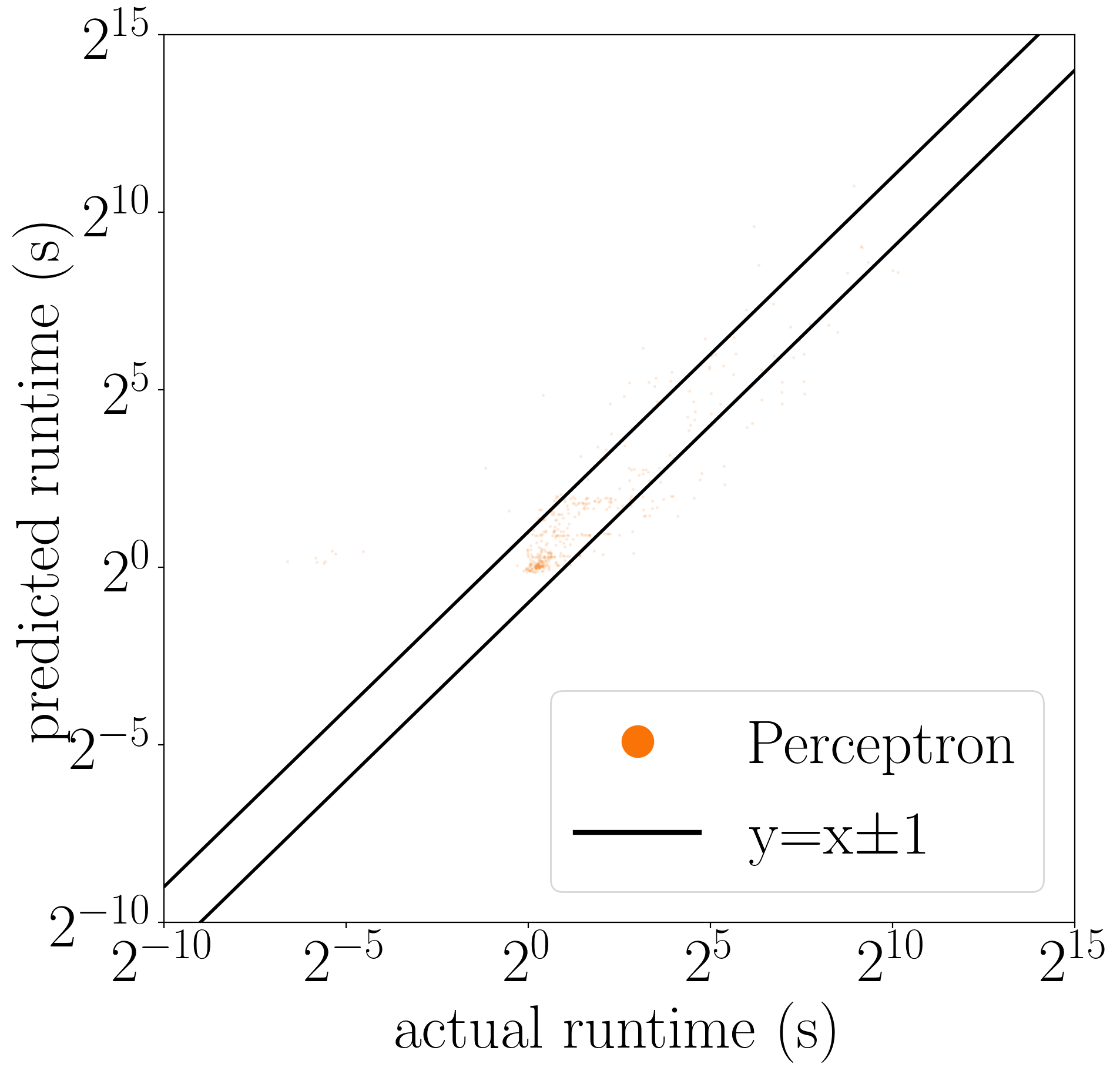}
		\end{subfigure}%
		\begin{subfigure}[b]{0.33\linewidth}
			\includegraphics[width=\linewidth]{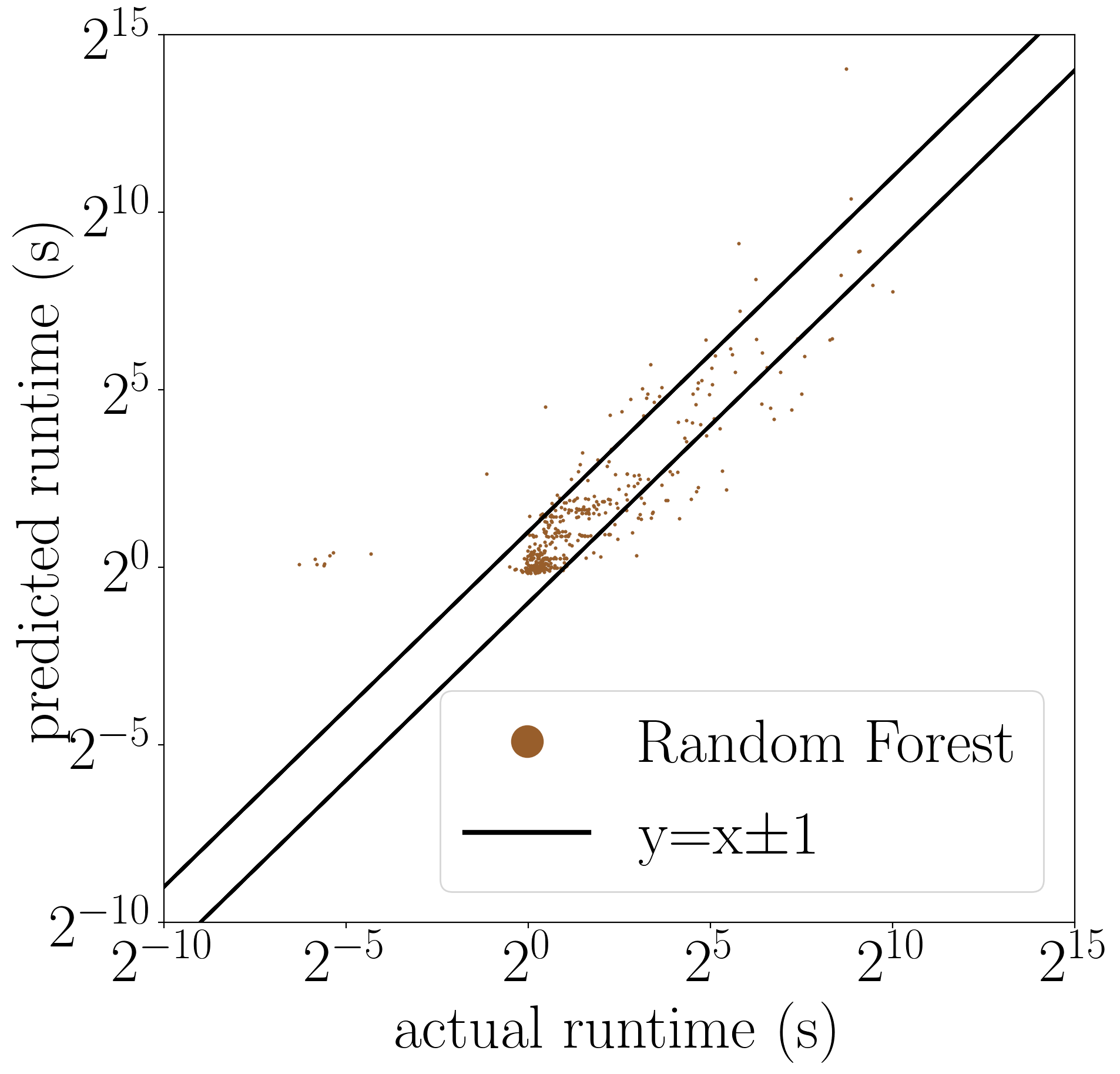}
		\end{subfigure}%

		\caption{Runtime prediction performance on different machine learning algorithms, on midsize OpenML datasets.}
		\label{fig:runtime_prediction_by_type}
	\end{figure*}

\textsc{Oboe} performs well in comparison with other AutoML methods
despite making a rather strong assumption about the structure of
model performance across datasets: namely, bilinearity.
It also requires effective predictions for model runtime.
In this section, we perform additional experiments on components of the
\textsc{Oboe} system to elucidate why the method works,
whether our assumptions are warranted,
and how they depend on detailed modeling choices.

\myparagraph{Low rank under different metrics}
\textsc{Oboe} uses balanced error rate to construct the error matrix, and
works on the premise that the error matrix can be approximated by a low rank matrix.
However, there is nothing special about the balanced error rate metric: most metrics result in an approximately low rank error matrix.
For example, when using the AUC metric to measure error,
the 418-by-219 error matrix from midsize OpenML datasets
has only 38 eigenvalues greater than 1\% of the largest, and 12 greater than 3\%.

\myparagraph{(Nonnegative) low rank structure of the error matrix}\label{section:nmf}
The features computed by PCA are dense and in general difficult to interpret.
In contrast, nonnegative matrix factorization (NMF) produces sparse positive feature vectors
and is thus widely used for clustering and interpretability \cite{xu2003document, kim2008sparse, turkmen2015review}.
We perform NMF on the error matrix $E$ to find nonnegative factors
$ W \in \mathbb{R}^{m \times k} $ and $ H \in \mathbb{R}^{k \times n}$
so that $ E \approx WH $.
Cluster membership of each model is given by the largest entry in its corresponding column in $ H $.

\begin{figure}
\centering
\includegraphics[width=.75\linewidth]{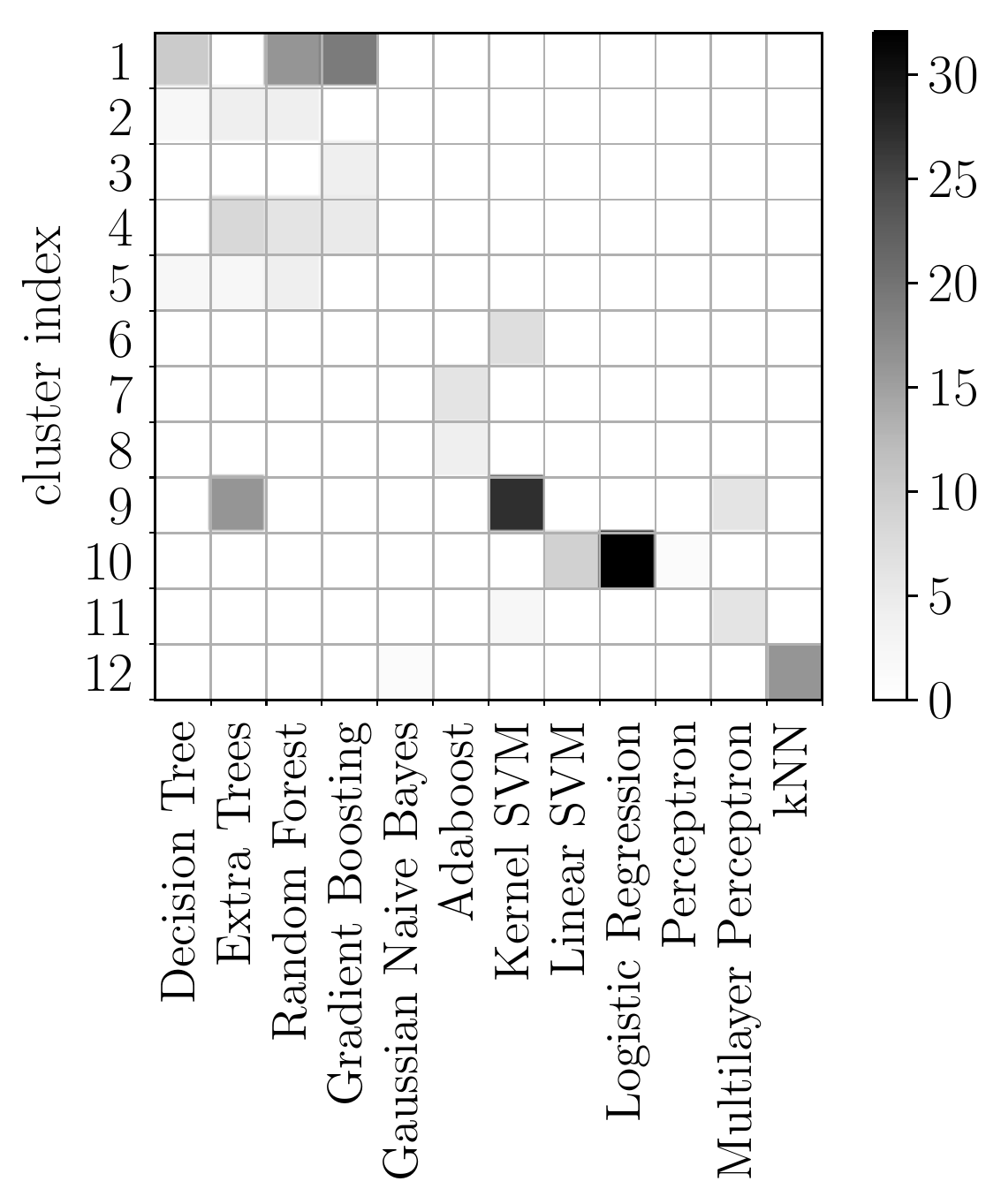}
\caption{Algorithm heatmap in clusters.
	Each block is colored by the number of models of the corresponding algorithm type in that cluster.
	Numbers next to the scale bar refer to the numbers of models.
}
\label{fig:nmf}
\end{figure}

Figure~\ref{fig:nmf} shows the heatmap of algorithms in clusters when $ k = 12 $
(the number of singular values no smaller than 3\% of the largest one).
Algorithm types are sparse in clusters: each cluster contains at most 3 types of algorithm.
Also, models belonging to the same kinds of algorithms tend to aggregate into the same clusters:
for example, Clusters 1 and 4 mainly consist of tree-based models;
Cluster 10 of linear models;
and Cluster 12 of neighborhood models.

\myparagraph{Runtime prediction performance}
\label{section:runtime_prediction}
Runtimes of linear models are among the most difficult to predict,
since they depend strongly on the conditioning of the problem.
Our runtime prediction accuracy on midsize OpenML datasets is shown in Table~\ref{table:accuracy} and in Figure~\ref{fig:runtime_prediction_by_type}.
We can see that our empirical prediction of model runtime is roughly unbiased.
Thus the sum of predicted runtimes on multiple models is a roughly good estimate.

\begin{table}
	\captionof{table}{Runtime prediction accuracy on OpenML datasets} 
	\label{table:accuracy}
	\centering
	\begin{tabular}{lrr}
		\hline
		Algorithm type    &   \multicolumn{2}{c}{Runtime prediction accuracy} \\
		& within factor of 2 & within factor of 4\\
		\hline
		Adaboost & 83.6\% & 94.3\%\\
		Decision tree & 76.7\% & 88.1\%\\
		Extra trees	& 96.6\% & 99.5\%\\
		Gradient boosting & 53.9\% & 84.3\%\\
		Gaussian naive Bayes & 89.6\% & 96.7\%\\
		kNN	& 85.2\% & 88.2\%\\
		Logistic regression	& 41.1\% & 76.0\%\\
		Multilayer perceptron & 78.9\% & 96.0\%\\
		Perceptron & 75.4\% & 94.3\%\\
		Random Forest & 94.4\% & 98.2\%\\
		Kernel SVM & 59.9\% & 86.7\%\\
		Linear SVM & 30.1\% & 73.2\%\\

		\hline
	\end{tabular}

\end{table}

\begin{figure}
\centering
\begin{minipage}[b]{.47\linewidth}
\includegraphics[width=\linewidth]{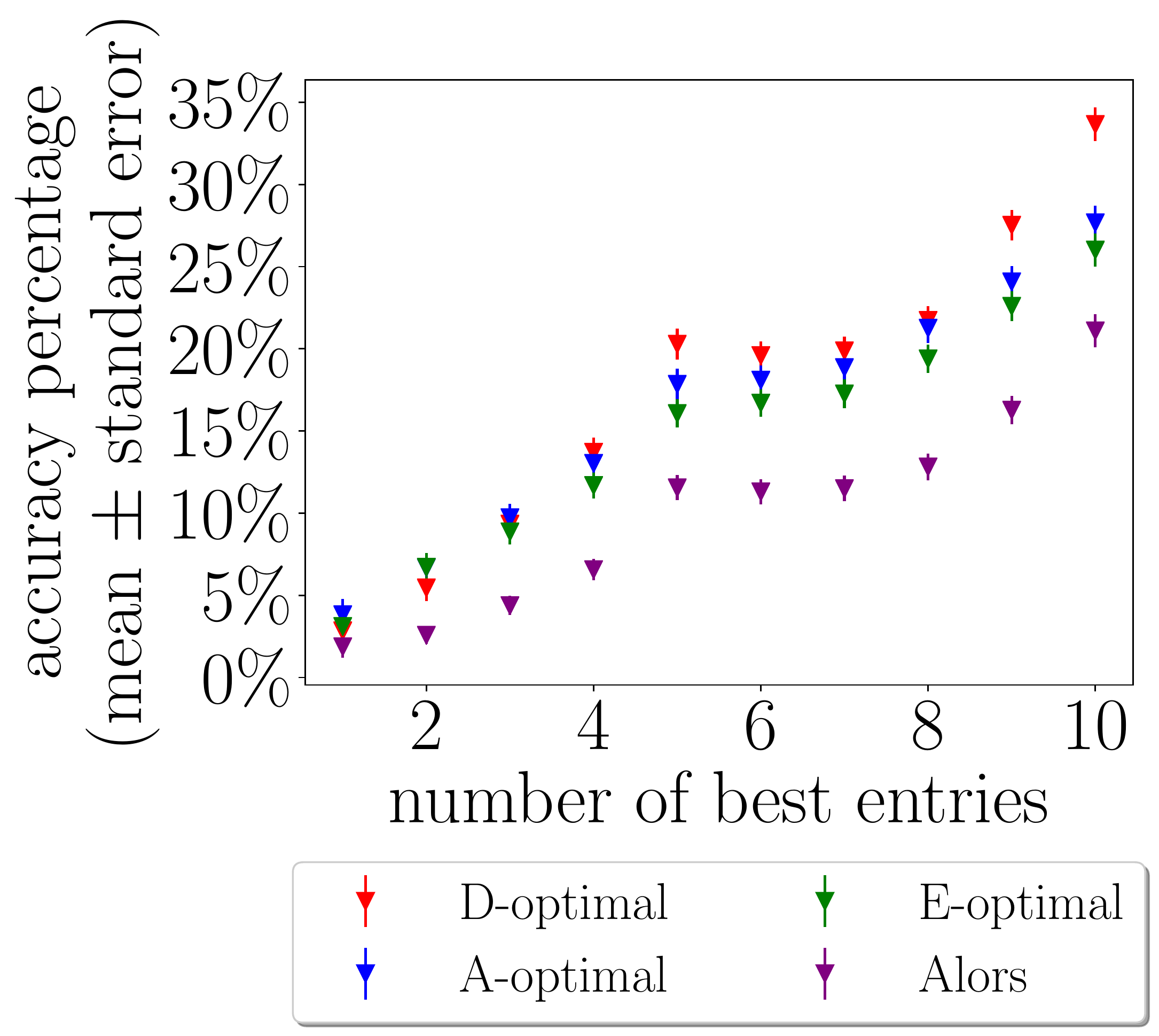}
\caption{Comparison of cold-start methods.
\newline}
\label{fig:cold_start}
\end{minipage}\hspace{0.05\linewidth}
\begin{minipage}[b]{.47\linewidth}
\includegraphics[width=\linewidth]{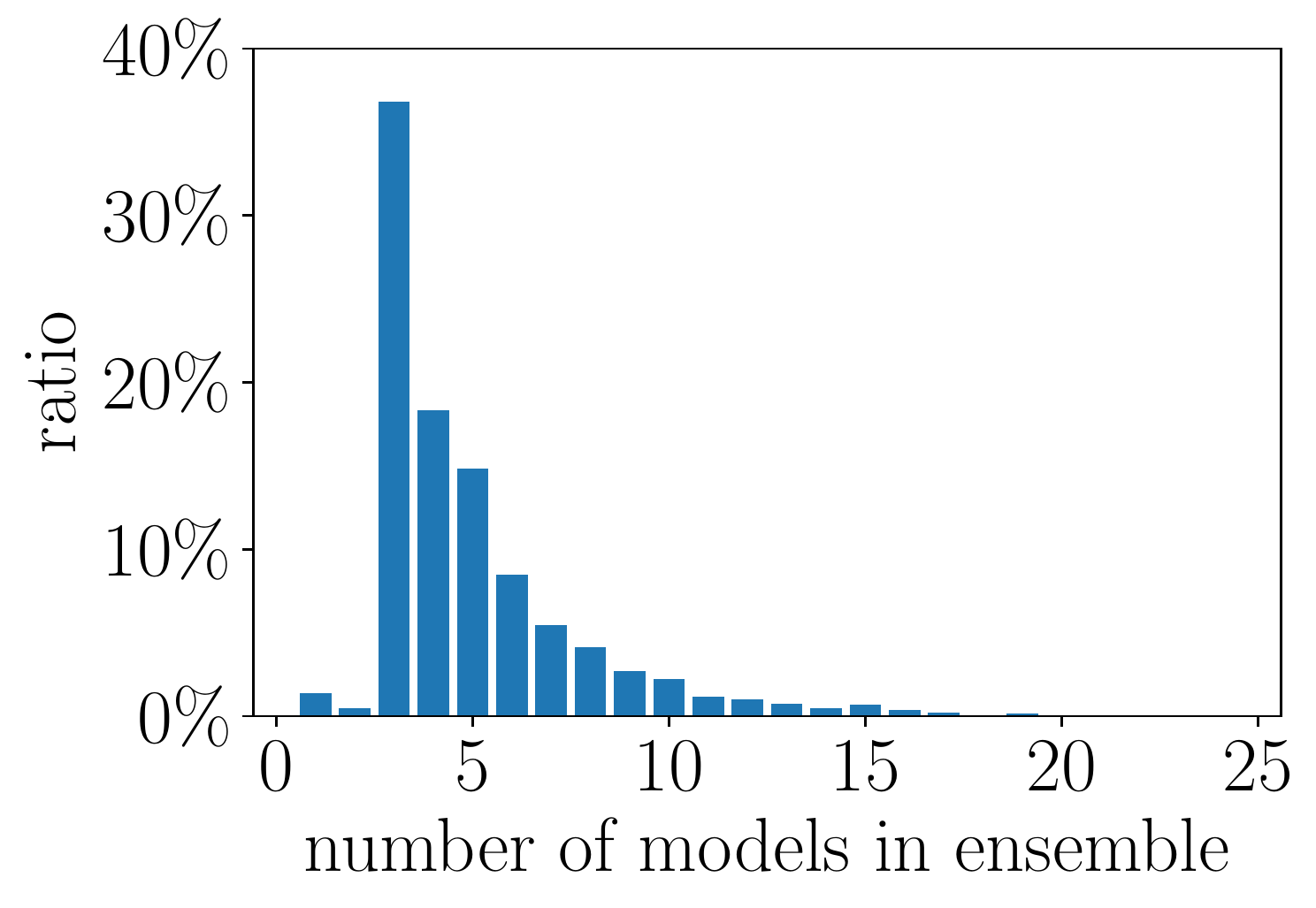}
\caption{Histogram of \textsc{Oboe} ensemble size. The ensembles were built in executions on midsize OpenML datasets in Section~\ref{comparison_with_auto_sklearn}.}
\label{fig:ensemble_size}
\end{minipage}
\end{figure}

\myparagraph{Cold-start}
\textsc{Oboe} uses D-optimal experiment design to cold-start model selection.
In Figure~\ref{fig:cold_start},
we compare this choice with A- and E-optimal design and nonlinear regression in Alors \cite{misir2017alors},
by means of leave-one-out cross-validation on midsize OpenML datasets.
We measure performance by the relative RMSE $\|e - \hat e\|_2 / \|e\|_2$
of the predicted performance vector
and by the number of correctly predicted best models, both averaged across datasets.
The approximate rank of the error matrix is set to be the
number of eigenvalues larger than 1\% of the largest, which is 38 here.
The time limit in experiment design implementation is set to be 4 seconds;
the nonlinear regressor used in Alors implementation is the default \texttt{RandomForestRegressor} in scikit-learn 0.19.2 \cite{scikit-learn}.

The horizontal axis is the number of models selected;
the vertical axis is the percentage of
best-ranked models shared between true and predicted performance vectors.
D-optimal design robustly outperforms.

\myparagraph{Ensemble size}
As shown in Figure~\ref{fig:ensemble_size}, more than 70\% of the ensembles constructed on midsize OpenML datasets have no more than 5 base learners.
This parsimony makes our ensembles easy to implement and interpret.

\section{Summary}
\textsc{Oboe} is an AutoML system that uses
collaborative filtering and optimal experiment design to predict performance of machine learning models.
By fitting a few models on the meta-test dataset,
this system transfers knowledge from meta-training datasets to select a promising set of models.
\textsc{Oboe} naturally handles different algorithm and hyperparameter types and can match state-of-the-art performance of AutoML systems much more quickly than competing approaches.

This work demonstrates the promise of collaborative filtering approaches to AutoML.
However, there is much more left to do.
Future work is needed to adapt \textsc{Oboe} to different loss metrics, budget types,
sparsely observed error matrices, and a wider range of machine learning algorithms.
Adapting a collaborative filtering approach to search for good machine learning
\emph{pipelines}, rather than individual algorithms, presents a more substantial challenge.
We also hope to see more approaches to the challenge of choosing hyper-hyperparameter settings
subject to limited computation and data: meta-learning is generally data(set)-constrained.
With continuing efforts by the AutoML community, we look forward to a world in which
domain experts seeking to use machine learning can focus on data
quality and problem formulation, rather than on tasks --- such as
algorithm selection and hyperparameter tuning ---
which are suitable for automation.

%
\begin{acks}
This work was supported in part by DARPA Award FA8750-17-2-0101. The authors thank Christophe Giraud-Carrier, Ameet Talwalkar, Raul Astudillo Marban, Matthew Zalesak, Lijun Ding and Davis Wertheimer for helpful discussions, thank Jack Dunn for a script to parse UCI Machine Learning Repository datasets, and also thank several anonymous reviewers for useful comments. 
\end{acks}

%
\bibliographystyle{ACM-Reference-Format}
\bibliography{main}


\begin{thebibliography}{46}


\ifx \showCODEN    \undefined \def \showCODEN     #1{\unskip}     \fi
\ifx \showDOI      \undefined \def \showDOI       #1{#1}\fi
\ifx \showISBNx    \undefined \def \showISBNx     #1{\unskip}     \fi
\ifx \showISBNxiii \undefined \def \showISBNxiii  #1{\unskip}     \fi
\ifx \showISSN     \undefined \def \showISSN      #1{\unskip}     \fi
\ifx \showLCCN     \undefined \def \showLCCN      #1{\unskip}     \fi
\ifx \shownote     \undefined \def \shownote      #1{#1}          \fi
\ifx \showarticletitle \undefined \def \showarticletitle #1{#1}   \fi
\ifx \showURL      \undefined \def \showURL       {\relax}        \fi
\providecommand\bibfield[2]{#2}
\providecommand\bibinfo[2]{#2}
\providecommand\natexlab[1]{#1}
\providecommand\showeprint[2][]{arXiv:#2}

\bibitem[\protect\citeauthoryear{Bardenet, Brendel, K{\'e}gl, and
  Sebag}{Bardenet et~al\mbox{.}}{2013}]%
        {bardenet2013collaborative}
\bibfield{author}{\bibinfo{person}{R{\'e}mi Bardenet},
  \bibinfo{person}{M{\'a}ty{\'a}s Brendel}, \bibinfo{person}{Bal{\'a}zs
  K{\'e}gl}, {and} \bibinfo{person}{Michele Sebag}.}
  \bibinfo{year}{2013}\natexlab{}.
\newblock \showarticletitle{{Collaborative hyperparameter tuning}}. In
  \bibinfo{booktitle}{\emph{ICML}}. \bibinfo{pages}{199--207}.
\newblock


\bibitem[\protect\citeauthoryear{Bartz-Beielstein and Markon}{Bartz-Beielstein
  and Markon}{2004}]%
        {bartz2004tuning}
\bibfield{author}{\bibinfo{person}{Thomas Bartz-Beielstein} {and}
  \bibinfo{person}{Sandor Markon}.} \bibinfo{year}{2004}\natexlab{}.
\newblock \showarticletitle{{Tuning search algorithms for real-world
  applications: A regression tree based approach}}. In
  \bibinfo{booktitle}{\emph{Congress on Evolutionary Computation}},
  Vol.~\bibinfo{volume}{1}. IEEE, \bibinfo{pages}{1111--1118}.
\newblock


\bibitem[\protect\citeauthoryear{Bergstra, Bardenet, Bengio, and
  K{\'e}gl}{Bergstra et~al\mbox{.}}{2011}]%
        {bergstra2011algorithms}
\bibfield{author}{\bibinfo{person}{James~S Bergstra}, \bibinfo{person}{R{\'e}mi
  Bardenet}, \bibinfo{person}{Yoshua Bengio}, {and} \bibinfo{person}{Bal{\'a}zs
  K{\'e}gl}.} \bibinfo{year}{2011}\natexlab{}.
\newblock \showarticletitle{{Algorithms for hyper-parameter optimization}}. In
  \bibinfo{booktitle}{\emph{Advances in Neural Information Processing
  Systems}}. \bibinfo{pages}{2546--2554}.
\newblock


\bibitem[\protect\citeauthoryear{Bischl, Richter, Bossek, Horn, Thomas, and
  Lang}{Bischl et~al\mbox{.}}{2017}]%
        {bischl2017mlrmbo}
\bibfield{author}{\bibinfo{person}{Bernd Bischl}, \bibinfo{person}{Jakob
  Richter}, \bibinfo{person}{Jakob Bossek}, \bibinfo{person}{Daniel Horn},
  \bibinfo{person}{Janek Thomas}, {and} \bibinfo{person}{Michel Lang}.}
  \bibinfo{year}{2017}\natexlab{}.
\newblock \showarticletitle{{mlrMBO: A modular framework for model-based
  optimization of expensive black-box functions}}.
\newblock \bibinfo{journal}{\emph{arXiv preprint arXiv:1703.03373}}
  (\bibinfo{year}{2017}).
\newblock


\bibitem[\protect\citeauthoryear{Boyd and Vandenberghe}{Boyd and
  Vandenberghe}{2004}]%
        {boyd2004convex}
\bibfield{author}{\bibinfo{person}{Stephen Boyd} {and} \bibinfo{person}{Lieven
  Vandenberghe}.} \bibinfo{year}{2004}\natexlab{}.
\newblock \bibinfo{booktitle}{\emph{{Convex optimization}}}.
\newblock \bibinfo{publisher}{Cambridge University Press}.
\newblock


\bibitem[\protect\citeauthoryear{Caruana, Munson, and Niculescu-Mizil}{Caruana
  et~al\mbox{.}}{2006}]%
        {caruana2006getting}
\bibfield{author}{\bibinfo{person}{Rich Caruana}, \bibinfo{person}{Art Munson},
  {and} \bibinfo{person}{Alexandru Niculescu-Mizil}.}
  \bibinfo{year}{2006}\natexlab{}.
\newblock \showarticletitle{{Getting the most out of ensemble selection}}. In
  \bibinfo{booktitle}{\emph{ICDM}}. IEEE, \bibinfo{pages}{828--833}.
\newblock


\bibitem[\protect\citeauthoryear{Caruana, Niculescu-Mizil, Crew, and
  Ksikes}{Caruana et~al\mbox{.}}{2004}]%
        {caruana2004ensemble}
\bibfield{author}{\bibinfo{person}{Rich Caruana}, \bibinfo{person}{Alexandru
  Niculescu-Mizil}, \bibinfo{person}{Geoff Crew}, {and} \bibinfo{person}{Alex
  Ksikes}.} \bibinfo{year}{2004}\natexlab{}.
\newblock \showarticletitle{{Ensemble selection from libraries of models}}. In
  \bibinfo{booktitle}{\emph{ICML}}. ACM, \bibinfo{pages}{18}.
\newblock


\bibitem[\protect\citeauthoryear{Chen, Wu, Mo, Chattopadhyay, and Lipson}{Chen
  et~al\mbox{.}}{2018}]%
        {chen2018autostacker}
\bibfield{author}{\bibinfo{person}{Boyuan Chen}, \bibinfo{person}{Harvey Wu},
  \bibinfo{person}{Warren Mo}, \bibinfo{person}{Ishanu Chattopadhyay}, {and}
  \bibinfo{person}{Hod Lipson}.} \bibinfo{year}{2018}\natexlab{}.
\newblock \showarticletitle{Autostacker: A compositional evolutionary learning
  system}. In \bibinfo{booktitle}{\emph{Proceedings of the Genetic and
  Evolutionary Computation Conference}}. ACM, \bibinfo{pages}{402--409}.
\newblock


\bibitem[\protect\citeauthoryear{Cunha, Soares, and de~Carvalho}{Cunha
  et~al\mbox{.}}{2018}]%
        {Cunha:2018:CRC:3240323.3240378}
\bibfield{author}{\bibinfo{person}{Tiago Cunha}, \bibinfo{person}{Carlos
  Soares}, {and} \bibinfo{person}{Andr{\'e} C. P. L.~F. de Carvalho}.}
  \bibinfo{year}{2018}\natexlab{}.
\newblock \showarticletitle{CF4CF: Recommending Collaborative Filtering
  Algorithms Using Collaborative Filtering}. In
  \bibinfo{booktitle}{\emph{Proceedings of the 12th ACM Conference on
  Recommender Systems}} \emph{(\bibinfo{series}{RecSys '18})}.
  \bibinfo{publisher}{ACM}, \bibinfo{address}{New York, NY, USA},
  \bibinfo{pages}{357--361}.
\newblock
\showISBNx{978-1-4503-5901-6}
\urldef\tempurl%
\url{https://doi.org/10.1145/3240323.3240378}
\showDOI{\tempurl}


\bibitem[\protect\citeauthoryear{Dheeru and Karra~Taniskidou}{Dheeru and
  Karra~Taniskidou}{2017}]%
        {Dua:2017}
\bibfield{author}{\bibinfo{person}{Dua Dheeru} {and} \bibinfo{person}{Efi
  Karra~Taniskidou}.} \bibinfo{year}{2017}\natexlab{}.
\newblock \bibinfo{title}{{UCI} Machine Learning Repository}.
\newblock
\newblock
\urldef\tempurl%
\url{http://archive.ics.uci.edu/ml}
\showURL{%
\tempurl}


\bibitem[\protect\citeauthoryear{Drori, Krishnamurthy, Rampin,
  de~Paula~Lourenco, Ono, Cho, Silva, and Freire}{Drori et~al\mbox{.}}{2018}]%
        {drori2018alphad3m}
\bibfield{author}{\bibinfo{person}{Iddo Drori}, \bibinfo{person}{Yamuna
  Krishnamurthy}, \bibinfo{person}{Remi Rampin}, \bibinfo{person}{Raoni de
  Paula~Lourenco}, \bibinfo{person}{Jorge~Piazentin Ono},
  \bibinfo{person}{Kyunghyun Cho}, \bibinfo{person}{Claudio Silva}, {and}
  \bibinfo{person}{Juliana Freire}.} \bibinfo{year}{2018}\natexlab{}.
\newblock \showarticletitle{AlphaD3M: Machine learning pipeline synthesis}. In
  \bibinfo{booktitle}{\emph{AutoML Workshop at ICML}}.
\newblock


\bibitem[\protect\citeauthoryear{Feurer, Klein, Eggensperger, Springenberg,
  Blum, and Hutter}{Feurer et~al\mbox{.}}{2015}]%
        {feurer2015efficient}
\bibfield{author}{\bibinfo{person}{Matthias Feurer}, \bibinfo{person}{Aaron
  Klein}, \bibinfo{person}{Katharina Eggensperger}, \bibinfo{person}{Jost
  Springenberg}, \bibinfo{person}{Manuel Blum}, {and} \bibinfo{person}{Frank
  Hutter}.} \bibinfo{year}{2015}\natexlab{}.
\newblock \showarticletitle{{Efficient and robust automated machine learning}}.
  In \bibinfo{booktitle}{\emph{Advances in Neural Information Processing
  Systems}}. \bibinfo{pages}{2962--2970}.
\newblock


\bibitem[\protect\citeauthoryear{Feurer, Springenberg, and Hutter}{Feurer
  et~al\mbox{.}}{2014}]%
        {feurer2014using}
\bibfield{author}{\bibinfo{person}{Matthias Feurer},
  \bibinfo{person}{Jost~Tobias Springenberg}, {and} \bibinfo{person}{Frank
  Hutter}.} \bibinfo{year}{2014}\natexlab{}.
\newblock \showarticletitle{{Using meta-learning to initialize Bayesian
  optimization of hyperparameters}}. In \bibinfo{booktitle}{\emph{International
  Conference on Meta-learning and Algorithm Selection}}. Citeseer,
  \bibinfo{pages}{3--10}.
\newblock


\bibitem[\protect\citeauthoryear{Fusi, Sheth, and Elibol}{Fusi
  et~al\mbox{.}}{2018}]%
        {fusi2018probabilistic}
\bibfield{author}{\bibinfo{person}{Nicolo Fusi}, \bibinfo{person}{Rishit
  Sheth}, {and} \bibinfo{person}{Melih Elibol}.}
  \bibinfo{year}{2018}\natexlab{}.
\newblock \showarticletitle{Probabilistic matrix factorization for automated
  machine learning}. In \bibinfo{booktitle}{\emph{Advances in Neural
  Information Processing Systems}}. \bibinfo{pages}{3352--3361}.
\newblock


\bibitem[\protect\citeauthoryear{Golub and Van~Loan}{Golub and
  Van~Loan}{2012}]%
        {golub2012matrix}
\bibfield{author}{\bibinfo{person}{Gene~H Golub} {and}
  \bibinfo{person}{Charles~F Van~Loan}.} \bibinfo{year}{2012}\natexlab{}.
\newblock \bibinfo{booktitle}{\emph{{Matrix computations}}}.
\newblock \bibinfo{publisher}{JHU Press}.
\newblock


\bibitem[\protect\citeauthoryear{Hazan, Klivans, and Yuan}{Hazan
  et~al\mbox{.}}{2018}]%
        {hazan2018hyperparameter}
\bibfield{author}{\bibinfo{person}{Elad Hazan}, \bibinfo{person}{Adam Klivans},
  {and} \bibinfo{person}{Yang Yuan}.} \bibinfo{year}{2018}\natexlab{}.
\newblock \showarticletitle{Hyperparameter optimization: a spectral approach}.
  In \bibinfo{booktitle}{\emph{ICLR}}.
\newblock
\urldef\tempurl%
\url{https://openreview.net/forum?id=H1zriGeCZ}
\showURL{%
\tempurl}


\bibitem[\protect\citeauthoryear{Herbrich, Lawrence, and Seeger}{Herbrich
  et~al\mbox{.}}{2003}]%
        {herbrich2003fast}
\bibfield{author}{\bibinfo{person}{Ralf Herbrich}, \bibinfo{person}{Neil~D
  Lawrence}, {and} \bibinfo{person}{Matthias Seeger}.}
  \bibinfo{year}{2003}\natexlab{}.
\newblock \showarticletitle{{Fast sparse Gaussian process methods: The
  informative vector machine}}. In \bibinfo{booktitle}{\emph{Advances in Neural
  Information Processing Systems}}. \bibinfo{pages}{625--632}.
\newblock


\bibitem[\protect\citeauthoryear{Huang, Jia, Yu, Chun, Maniatis, and
  Naik}{Huang et~al\mbox{.}}{2010}]%
        {huang2010predicting}
\bibfield{author}{\bibinfo{person}{Ling Huang}, \bibinfo{person}{Jinzhu Jia},
  \bibinfo{person}{Bin Yu}, \bibinfo{person}{Byung-Gon Chun},
  \bibinfo{person}{Petros Maniatis}, {and} \bibinfo{person}{Mayur Naik}.}
  \bibinfo{year}{2010}\natexlab{}.
\newblock \showarticletitle{{Predicting execution time of computer programs
  using sparse polynomial regression}}. In \bibinfo{booktitle}{\emph{Advances
  in Neural Information Processing Systems}}. \bibinfo{pages}{883--891}.
\newblock


\bibitem[\protect\citeauthoryear{Hutter, Hamadi, Hoos, and Leyton-Brown}{Hutter
  et~al\mbox{.}}{2006}]%
        {hutter2006performance}
\bibfield{author}{\bibinfo{person}{Frank Hutter}, \bibinfo{person}{Youssef
  Hamadi}, \bibinfo{person}{Holger~H Hoos}, {and} \bibinfo{person}{Kevin
  Leyton-Brown}.} \bibinfo{year}{2006}\natexlab{}.
\newblock \showarticletitle{{Performance prediction and automated tuning of
  randomized and parametric algorithms}}. In
  \bibinfo{booktitle}{\emph{International Conference on Principles and Practice
  of Constraint Programming}}. Springer, \bibinfo{pages}{213--228}.
\newblock


\bibitem[\protect\citeauthoryear{Hutter, Hoos, and Leyton-Brown}{Hutter
  et~al\mbox{.}}{2011}]%
        {hutter2011sequential}
\bibfield{author}{\bibinfo{person}{Frank Hutter}, \bibinfo{person}{Holger~H
  Hoos}, {and} \bibinfo{person}{Kevin Leyton-Brown}.}
  \bibinfo{year}{2011}\natexlab{}.
\newblock \showarticletitle{{Sequential Model-Based Optimization for General
  Algorithm Configuration.}}
\newblock \bibinfo{journal}{\emph{LION}}  \bibinfo{volume}{5}
  (\bibinfo{year}{2011}), \bibinfo{pages}{507--523}.
\newblock


\bibitem[\protect\citeauthoryear{Hutter, Xu, Hoos, and Leyton-Brown}{Hutter
  et~al\mbox{.}}{2014}]%
        {hutter2014algorithm}
\bibfield{author}{\bibinfo{person}{Frank Hutter}, \bibinfo{person}{Lin Xu},
  \bibinfo{person}{Holger~H Hoos}, {and} \bibinfo{person}{Kevin Leyton-Brown}.}
  \bibinfo{year}{2014}\natexlab{}.
\newblock \showarticletitle{{Algorithm runtime prediction: Methods \&
  evaluation}}.
\newblock \bibinfo{journal}{\emph{Artificial Intelligence}}
  \bibinfo{volume}{206} (\bibinfo{year}{2014}), \bibinfo{pages}{79--111}.
\newblock


\bibitem[\protect\citeauthoryear{John and Draper}{John and Draper}{1975}]%
        {john1975d}
\bibfield{author}{\bibinfo{person}{RC~St John} {and} \bibinfo{person}{Norman~R
  Draper}.} \bibinfo{year}{1975}\natexlab{}.
\newblock \showarticletitle{{D-optimality for regression designs: a review}}.
\newblock \bibinfo{journal}{\emph{Technometrics}} \bibinfo{volume}{17},
  \bibinfo{number}{1} (\bibinfo{year}{1975}), \bibinfo{pages}{15--23}.
\newblock


\bibitem[\protect\citeauthoryear{Kim and Park}{Kim and Park}{2008}]%
        {kim2008sparse}
\bibfield{author}{\bibinfo{person}{Jingu Kim} {and} \bibinfo{person}{Haesun
  Park}.} \bibinfo{year}{2008}\natexlab{}.
\newblock \bibinfo{booktitle}{\emph{Sparse nonnegative matrix factorization for
  clustering}}.
\newblock \bibinfo{type}{{T}echnical {R}eport}. \bibinfo{institution}{Georgia
  Institute of Technology}.
\newblock


\bibitem[\protect\citeauthoryear{Krause, Singh, and Guestrin}{Krause
  et~al\mbox{.}}{2008}]%
        {krause2008near}
\bibfield{author}{\bibinfo{person}{Andreas Krause}, \bibinfo{person}{Ajit
  Singh}, {and} \bibinfo{person}{Carlos Guestrin}.}
  \bibinfo{year}{2008}\natexlab{}.
\newblock \showarticletitle{{Near-optimal sensor placements in Gaussian
  processes: Theory, efficient algorithms and empirical studies}}.
\newblock \bibinfo{journal}{\emph{Journal of Machine Learning Research}}
  \bibinfo{volume}{9}, \bibinfo{number}{Feb} (\bibinfo{year}{2008}),
  \bibinfo{pages}{235--284}.
\newblock


\bibitem[\protect\citeauthoryear{Leite, Brazdil, and Vanschoren}{Leite
  et~al\mbox{.}}{2012}]%
        {leite2012selecting}
\bibfield{author}{\bibinfo{person}{Rui Leite}, \bibinfo{person}{Pavel Brazdil},
  {and} \bibinfo{person}{Joaquin Vanschoren}.} \bibinfo{year}{2012}\natexlab{}.
\newblock \showarticletitle{{Selecting classification algorithms with active
  testing}}. In \bibinfo{booktitle}{\emph{International Workshop on Machine
  Learning and Data Mining in Pattern Recognition}}. Springer,
  \bibinfo{pages}{117--131}.
\newblock


\bibitem[\protect\citeauthoryear{Lemke, Budka, and Gabrys}{Lemke
  et~al\mbox{.}}{2015}]%
        {lemke2015metalearning}
\bibfield{author}{\bibinfo{person}{Christiane Lemke}, \bibinfo{person}{Marcin
  Budka}, {and} \bibinfo{person}{Bogdan Gabrys}.}
  \bibinfo{year}{2015}\natexlab{}.
\newblock \showarticletitle{{Metalearning: a survey of trends and
  technologies}}.
\newblock \bibinfo{journal}{\emph{Artificial Intelligence Review}}
  \bibinfo{volume}{44}, \bibinfo{number}{1} (\bibinfo{year}{2015}),
  \bibinfo{pages}{117--130}.
\newblock


\bibitem[\protect\citeauthoryear{MacKay}{MacKay}{1992}]%
        {mackay1992information}
\bibfield{author}{\bibinfo{person}{David~JC MacKay}.}
  \bibinfo{year}{1992}\natexlab{}.
\newblock \showarticletitle{{Information-based objective functions for active
  data selection}}.
\newblock \bibinfo{journal}{\emph{Neural Computation}} \bibinfo{volume}{4},
  \bibinfo{number}{4} (\bibinfo{year}{1992}), \bibinfo{pages}{590--604}.
\newblock


\bibitem[\protect\citeauthoryear{M{\i}s{\i}r and Sebag}{M{\i}s{\i}r and
  Sebag}{2017}]%
        {misir2017alors}
\bibfield{author}{\bibinfo{person}{Mustafa M{\i}s{\i}r} {and}
  \bibinfo{person}{Mich{\`e}le Sebag}.} \bibinfo{year}{2017}\natexlab{}.
\newblock \showarticletitle{{Alors: An algorithm recommender system}}.
\newblock \bibinfo{journal}{\emph{Artificial Intelligence}}
  \bibinfo{volume}{244} (\bibinfo{year}{2017}), \bibinfo{pages}{291--314}.
\newblock


\bibitem[\protect\citeauthoryear{Mood et~al\mbox{.}}{Mood
  et~al\mbox{.}}{1946}]%
        {mood1946hotelling}
\bibfield{author}{\bibinfo{person}{Alexander~M Mood} {et~al\mbox{.}}}
  \bibinfo{year}{1946}\natexlab{}.
\newblock \showarticletitle{{On Hotelling's weighing problem}}.
\newblock \bibinfo{journal}{\emph{The Annals of Mathematical Statistics}}
  \bibinfo{volume}{17}, \bibinfo{number}{4} (\bibinfo{year}{1946}),
  \bibinfo{pages}{432--446}.
\newblock


\bibitem[\protect\citeauthoryear{Pedregosa, Varoquaux, Gramfort, Michel,
  Thirion, Grisel, Blondel, Prettenhofer, Weiss, Dubourg, Vanderplas, Passos,
  Cournapeau, Brucher, Perrot, and Duchesnay}{Pedregosa et~al\mbox{.}}{2011}]%
        {scikit-learn}
\bibfield{author}{\bibinfo{person}{F. Pedregosa}, \bibinfo{person}{G.
  Varoquaux}, \bibinfo{person}{A. Gramfort}, \bibinfo{person}{V. Michel},
  \bibinfo{person}{B. Thirion}, \bibinfo{person}{O. Grisel},
  \bibinfo{person}{M. Blondel}, \bibinfo{person}{P. Prettenhofer},
  \bibinfo{person}{R. Weiss}, \bibinfo{person}{V. Dubourg}, \bibinfo{person}{J.
  Vanderplas}, \bibinfo{person}{A. Passos}, \bibinfo{person}{D. Cournapeau},
  \bibinfo{person}{M. Brucher}, \bibinfo{person}{M. Perrot}, {and}
  \bibinfo{person}{E. Duchesnay}.} \bibinfo{year}{2011}\natexlab{}.
\newblock \showarticletitle{{Scikit-learn: Machine Learning in Python}}.
\newblock \bibinfo{journal}{\emph{Journal of Machine Learning Research}}
  \bibinfo{volume}{12} (\bibinfo{year}{2011}), \bibinfo{pages}{2825--2830}.
\newblock


\bibitem[\protect\citeauthoryear{Pfahringer, Bensusan, and
  Giraud-Carrier}{Pfahringer et~al\mbox{.}}{2000}]%
        {pfahringer2000meta}
\bibfield{author}{\bibinfo{person}{Bernhard Pfahringer}, \bibinfo{person}{Hilan
  Bensusan}, {and} \bibinfo{person}{Christophe~G Giraud-Carrier}.}
  \bibinfo{year}{2000}\natexlab{}.
\newblock \showarticletitle{{Meta-Learning by Landmarking Various Learning
  Algorithms}}. In \bibinfo{booktitle}{\emph{ICML}}. \bibinfo{pages}{743--750}.
\newblock


\bibitem[\protect\citeauthoryear{Pukelsheim}{Pukelsheim}{1993}]%
        {pukelsheim1993optimal}
\bibfield{author}{\bibinfo{person}{Friedrich Pukelsheim}.}
  \bibinfo{year}{1993}\natexlab{}.
\newblock \bibinfo{booktitle}{\emph{{Optimal design of experiments}}}.
  Vol.~\bibinfo{volume}{50}.
\newblock \bibinfo{publisher}{SIAM}.
\newblock


\bibitem[\protect\citeauthoryear{Rasmussen and Williams}{Rasmussen and
  Williams}{2006}]%
        {williams2006gaussian}
\bibfield{author}{\bibinfo{person}{Carl~Edward Rasmussen} {and}
  \bibinfo{person}{Christopher~KI Williams}.} \bibinfo{year}{2006}\natexlab{}.
\newblock \bibinfo{booktitle}{\emph{{Gaussian processes for machine
  learning}}}.
\newblock \bibinfo{publisher}{the MIT Press}.
\newblock


\bibitem[\protect\citeauthoryear{Sebastiani and Wynn}{Sebastiani and
  Wynn}{2000}]%
        {sebastiani2000maximum}
\bibfield{author}{\bibinfo{person}{Paola Sebastiani} {and}
  \bibinfo{person}{Henry~P Wynn}.} \bibinfo{year}{2000}\natexlab{}.
\newblock \showarticletitle{{Maximum entropy sampling and optimal Bayesian
  experimental design}}.
\newblock \bibinfo{journal}{\emph{Journal of the Royal Statistical Society:
  Series B (Statistical Methodology)}} \bibinfo{volume}{62},
  \bibinfo{number}{1} (\bibinfo{year}{2000}), \bibinfo{pages}{145--157}.
\newblock


\bibitem[\protect\citeauthoryear{Smith-Miles and van Hemert}{Smith-Miles and
  van Hemert}{2011}]%
        {smith2011discovering}
\bibfield{author}{\bibinfo{person}{Kate Smith-Miles} {and}
  \bibinfo{person}{Jano van Hemert}.} \bibinfo{year}{2011}\natexlab{}.
\newblock \showarticletitle{{Discovering the suitability of optimisation
  algorithms by learning from evolved instances}}.
\newblock \bibinfo{journal}{\emph{Annals of Mathematics and Artificial
  Intelligence}} \bibinfo{volume}{61}, \bibinfo{number}{2}
  (\bibinfo{year}{2011}), \bibinfo{pages}{87--104}.
\newblock


\bibitem[\protect\citeauthoryear{Snoek, Larochelle, and Adams}{Snoek
  et~al\mbox{.}}{2012}]%
        {snoek2012practical}
\bibfield{author}{\bibinfo{person}{Jasper Snoek}, \bibinfo{person}{Hugo
  Larochelle}, {and} \bibinfo{person}{Ryan~P Adams}.}
  \bibinfo{year}{2012}\natexlab{}.
\newblock \showarticletitle{{Practical Bayesian optimization of machine
  learning algorithms}}. In \bibinfo{booktitle}{\emph{Advances in Neural
  Information Processing Systems}}. \bibinfo{pages}{2951--2959}.
\newblock


\bibitem[\protect\citeauthoryear{Srinivas, Krause, Kakade, and Seeger}{Srinivas
  et~al\mbox{.}}{2010}]%
        {srinivas2009gaussian}
\bibfield{author}{\bibinfo{person}{Niranjan Srinivas}, \bibinfo{person}{Andreas
  Krause}, \bibinfo{person}{Sham Kakade}, {and} \bibinfo{person}{Matthias
  Seeger}.} \bibinfo{year}{2010}\natexlab{}.
\newblock \showarticletitle{{Gaussian Process Optimization in the Bandit
  Setting: No Regret and Experimental Design}}. In
  \bibinfo{booktitle}{\emph{ICML}}. \bibinfo{pages}{1015--1022}.
\newblock


\bibitem[\protect\citeauthoryear{Stern, Samulowitz, Herbrich, Graepel, Pulina,
  and Tacchella}{Stern et~al\mbox{.}}{2010}]%
        {stern2010collaborative}
\bibfield{author}{\bibinfo{person}{David~H Stern}, \bibinfo{person}{Horst
  Samulowitz}, \bibinfo{person}{Ralf Herbrich}, \bibinfo{person}{Thore
  Graepel}, \bibinfo{person}{Luca Pulina}, {and} \bibinfo{person}{Armando
  Tacchella}.} \bibinfo{year}{2010}\natexlab{}.
\newblock \showarticletitle{{Collaborative Expert Portfolio Management}}. In
  \bibinfo{booktitle}{\emph{AAAI}}. \bibinfo{pages}{179--184}.
\newblock


\bibitem[\protect\citeauthoryear{T{\"u}rkmen}{T{\"u}rkmen}{2015}]%
        {turkmen2015review}
\bibfield{author}{\bibinfo{person}{Ali~Caner T{\"u}rkmen}.}
  \bibinfo{year}{2015}\natexlab{}.
\newblock \showarticletitle{A review of nonnegative matrix factorization
  methods for clustering}.
\newblock \bibinfo{journal}{\emph{arXiv preprint arXiv:1507.03194}}
  (\bibinfo{year}{2015}).
\newblock


\bibitem[\protect\citeauthoryear{Udell and Townsend}{Udell and
  Townsend}{2019}]%
        {udell2019big}
\bibfield{author}{\bibinfo{person}{Madeleine Udell} {and} \bibinfo{person}{Alex
  Townsend}.} \bibinfo{year}{2019}\natexlab{}.
\newblock \showarticletitle{Why Are Big Data Matrices Approximately Low Rank?}
\newblock \bibinfo{journal}{\emph{SIAM Journal on Mathematics of Data Science}}
  \bibinfo{volume}{1}, \bibinfo{number}{1} (\bibinfo{year}{2019}),
  \bibinfo{pages}{144--160}.
\newblock


\bibitem[\protect\citeauthoryear{Vanschoren, van Rijn, Bischl, and
  Torgo}{Vanschoren et~al\mbox{.}}{2013}]%
        {OpenML2013}
\bibfield{author}{\bibinfo{person}{Joaquin Vanschoren}, \bibinfo{person}{Jan~N.
  van Rijn}, \bibinfo{person}{Bernd Bischl}, {and} \bibinfo{person}{Luis
  Torgo}.} \bibinfo{year}{2013}\natexlab{}.
\newblock \showarticletitle{{OpenML: Networked Science in Machine Learning}}.
\newblock \bibinfo{journal}{\emph{SIGKDD Explorations}} \bibinfo{volume}{15},
  \bibinfo{number}{2} (\bibinfo{year}{2013}), \bibinfo{pages}{49--60}.
\newblock
\urldef\tempurl%
\url{https://doi.org/10.1145/2641190.2641198}
\showDOI{\tempurl}


\bibitem[\protect\citeauthoryear{Wald}{Wald}{1943}]%
        {wald1943efficient}
\bibfield{author}{\bibinfo{person}{Abraham Wald}.}
  \bibinfo{year}{1943}\natexlab{}.
\newblock \showarticletitle{{On the efficient design of statistical
  investigations}}.
\newblock \bibinfo{journal}{\emph{The Annals of Mathematical Statistics}}
  \bibinfo{volume}{14}, \bibinfo{number}{2} (\bibinfo{year}{1943}),
  \bibinfo{pages}{134--140}.
\newblock


\bibitem[\protect\citeauthoryear{Wistuba, Schilling, and
  Schmidt-Thieme}{Wistuba et~al\mbox{.}}{2015}]%
        {wistuba2015learning}
\bibfield{author}{\bibinfo{person}{M. Wistuba}, \bibinfo{person}{N. Schilling},
  {and} \bibinfo{person}{L. Schmidt-Thieme}.} \bibinfo{year}{2015}\natexlab{}.
\newblock \showarticletitle{Learning hyperparameter optimization
  initializations}. In \bibinfo{booktitle}{\emph{IEEE International Conference
  on Data Science and Advanced Analytics}}. \bibinfo{pages}{1--10}.
\newblock
\urldef\tempurl%
\url{https://doi.org/10.1109/DSAA.2015.7344817}
\showDOI{\tempurl}


\bibitem[\protect\citeauthoryear{Xu, Liu, and Gong}{Xu et~al\mbox{.}}{2003}]%
        {xu2003document}
\bibfield{author}{\bibinfo{person}{Wei Xu}, \bibinfo{person}{Xin Liu}, {and}
  \bibinfo{person}{Yihong Gong}.} \bibinfo{year}{2003}\natexlab{}.
\newblock \showarticletitle{Document clustering based on non-negative matrix
  factorization}. In \bibinfo{booktitle}{\emph{Proceedings of the 26th annual
  international ACM SIGIR conference on Research and development in informaion
  retrieval}}. ACM, \bibinfo{pages}{267--273}.
\newblock


\bibitem[\protect\citeauthoryear{Yogatama and Mann}{Yogatama and Mann}{2014}]%
        {yogatama2014efficient}
\bibfield{author}{\bibinfo{person}{Dani Yogatama} {and} \bibinfo{person}{Gideon
  Mann}.} \bibinfo{year}{2014}\natexlab{}.
\newblock \showarticletitle{{Efficient transfer learning method for automatic
  hyperparameter tuning}}. In \bibinfo{booktitle}{\emph{Artificial Intelligence
  and Statistics}}. \bibinfo{pages}{1077--1085}.
\newblock


\bibitem[\protect\citeauthoryear{Zhang, Bahadori, Su, and Sun}{Zhang
  et~al\mbox{.}}{2016}]%
        {zhang2016flash}
\bibfield{author}{\bibinfo{person}{Yuyu Zhang}, \bibinfo{person}{Mohammad~Taha
  Bahadori}, \bibinfo{person}{Hang Su}, {and} \bibinfo{person}{Jimeng Sun}.}
  \bibinfo{year}{2016}\natexlab{}.
\newblock \showarticletitle{{FLASH: fast Bayesian optimization for data
  analytic pipelines}}. In \bibinfo{booktitle}{\emph{Proceedings of the 22nd
  ACM SIGKDD International Conference on Knowledge Discovery and Data Mining}}.
  ACM, \bibinfo{pages}{2065--2074}.
\newblock


\end{thebibliography}


%
\appendix
\vspace{20em}
\begin{table*}
	\caption{Base Algorithm and Hyperparameter Settings}
	\label{table:models}
	\centering
	\begin{tabular}{ll}
		\hline
		Algorithm type    &   Hyperparameter names (values) \\
		\hline
		Adaboost & \texttt{n\_estimators (50,100), learning\_rate (1.0,1.5,2.0,2.5,3)} \\
		Decision tree & \texttt{min\_samples\_split (2,4,8,16,32,64,128,256,512,1024,0.01,0.001,0.0001,1e-05)} \\
		Extra trees & \multicolumn{1}{p{8cm}}{\texttt{min\_samples\_split (2,4,8,16,32,64,128,256,512,1024,0.01,0.001,0.0001,1e-05), criterion (gini,entropy)}} \\
		Gradient boosting & \multicolumn{1}{p{13cm}}{\texttt{learning\_rate (0.001,0.01,0.025,0.05,0.1,0.25,0.5), max\_depth (3, 6), max\_features (null,log2)}} \\
		Gaussian naive Bayes & - \\
		kNN & \texttt{n\_neighbors (1,3,5,7,9,11,13,15), p (1,2)} \\
		Logistic regression & \texttt{C (0.25,0.5,0.75,1,1.5,2,3,4), solver (liblinear,saga), penalty (l1,l2)} \\
		Multilayer perceptron & \multicolumn{1}{p{13cm}}{\texttt{learning\_rate\_init (0.0001,0.001,0.01),  learning\_rate (adaptive), solver (sgd,adam), alpha (0.0001, 0.01)}}\\
		Perceptron & - \\
		Random forest & \multicolumn{1}{p{8cm}}{\texttt{min\_samples\_split (2,4,8,16,32,64,128,256,512,1024,0.01,0.001,0.0001,1e-05), criterion (gini,entropy)}} \\
		Kernel SVM & \texttt{C (0.125,0.25,0.5,0.75,1,2,4,8,16), kernel (rbf,poly), coef0 (0,10)} \\
		Linear SVM & \texttt{C (0.125,0.25,0.5,0.75,1,2,4,8,16)} \\
		\hline
	\end{tabular}
\end{table*}

For reproducibility, please refer to our GitHub repositories (the \textsc{Oboe} system: \url{https://github.com/udellgroup/oboe}; experiments: \url{https://github.com/udellgroup/oboe-testing}). Additional information is as follows.

\section{Machine Learning Models}\label{supp:models}
Shown in Table~\ref{table:models}, the hyperparameter names are the same as those in scikit-learn 0.19.2.

\section{Dataset meta-features} \label{supp:metafeature}
Dataset meta-features used throughout the experiments are listed in Table~\ref{table:metafeatures} (next page).

\begin{table*}
	\caption{Dataset Meta-features} 
	\label{table:metafeatures}
	\centering
	\begin{tabular}{lll}
		\hline
		Meta-feature name  &   Explanation \\
		\hline
		number of instances & number of data points in the dataset \\
		log number of instances &  \multicolumn{1}{p{6cm}}{the (natural) logarithm of number of instances}\\
		number of classes &  \\
		number of features &  \\
		log number of features & \multicolumn{1}{p{6cm}}{the (natural) logarithm of number of features} \\
		number of instances with missing values &  \\
		percentage of instances with missing values &  \\
		number of features with missing values & \\
		percentage of features with missing values &  \\
		number of missing values &  \\
		percentage of missing values & \\
		number of numeric features &  \\
		number of categorical features &  \\
		ratio numerical to nominal & the ratio of number of numerical features to the number of categorical features \\
		ratio numerical to nominal &  \\
		dataset ratio &  the ratio of number of features to the number of data points \\
		log dataset ratio &  the natural logarithm of dataset ratio \\
		inverse dataset ratio &  \\
		log inverse dataset ratio &  \\
		class probability (min, max, mean, std) &  the (min, max, mean, std) of ratios of data points in each class\\
		symbols (min, max, mean, std, sum) & the (min, max, mean, std, sum) of the numbers of symbols in all categorical features\\
		kurtosis (min, max, mean, std) &  \\
		skewness (min, max, mean, std) &  \\
		class entropy & the entropy of the distribution of class labels (logarithm base 2) \\
		& \\
		\textbf{landmarking \cite{pfahringer2000meta} meta-features} &  \\
		LDA & \\
		decision tree & decision tree classifier with 10-fold cross validation\\
		decision node learner & \multicolumn{1}{p{12cm}}{ 10-fold cross-validated decision tree classifier with \texttt{criterion="entropy", max\_depth=1, min\_samples\_split=2, min\_samples\_leaf=1,  max\_features=None}} \\
		random node learner & \multicolumn{1}{p{12cm}}{10-fold cross-validated decision tree classifier with \texttt{max\_features=1} and the same above for the rest}\\
		1-NN & \\
		PCA fraction of components for 95\% variance & the fraction of components that account for 95\% of variance\\
		PCA kurtosis first PC & kurtosis of the dimensionality-reduced data matrix along the first principal component\\
		PCA skewness first PC & skewness of the dimensionality-reduced data matrix along the first principal component\\
		
		\hline
	\end{tabular}
\end{table*}

\section{Meta-feature calculation time}
\label{metafeaturetime}
\begin{figure}[H]
	\begin{subfigure}[b]{.47\linewidth}
		\includegraphics[width=\linewidth]{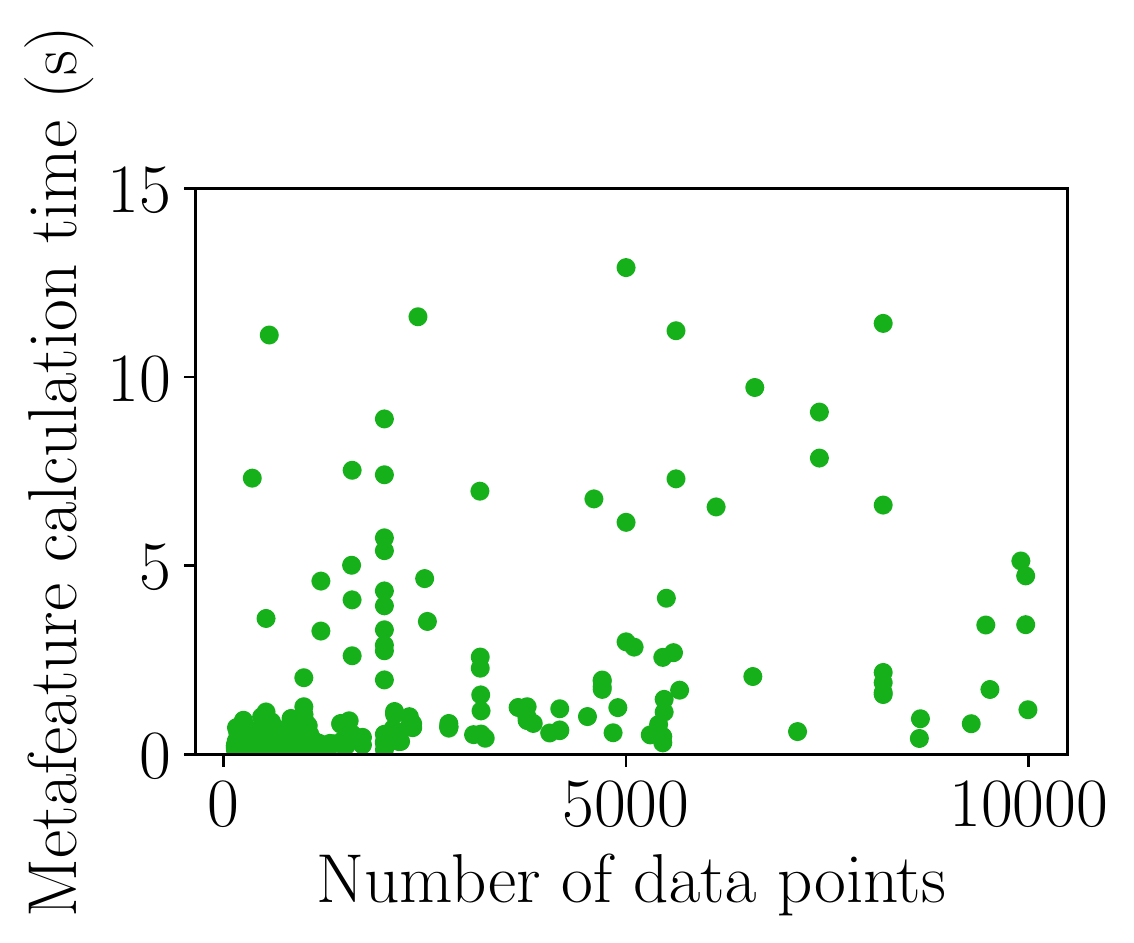}
		
	\end{subfigure}%
	\begin{subfigure}[b]{0.47\linewidth}
		\includegraphics[width=\linewidth]{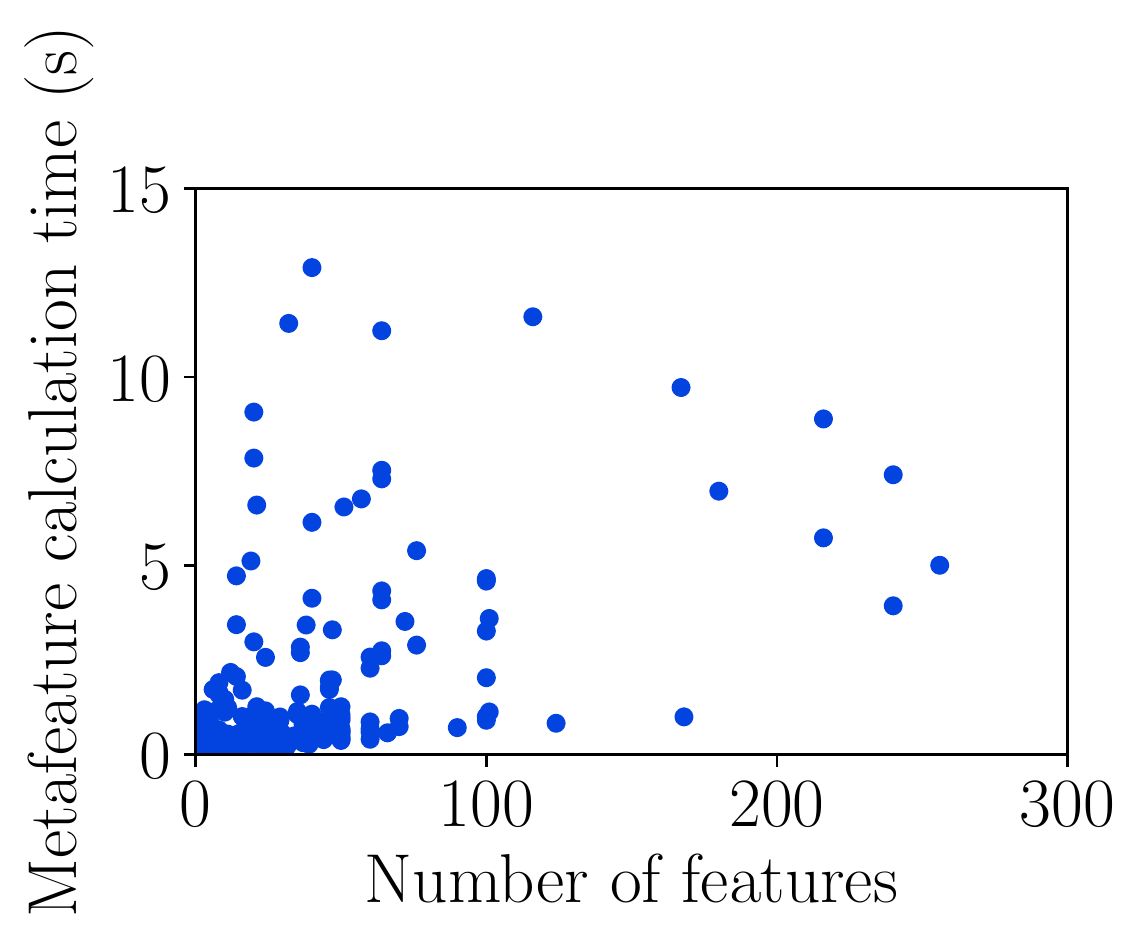}
		
	\end{subfigure}%
	
	\caption{Meta-feature calculation time and corresponding dataset sizes of the midsize OpenML datasets. The collection of meta-features is the same as that used by auto-sklearn \cite{feurer2015efficient}. We can see some calculation times are not negligible.}
	
	\label{fig:metafeature_calculation}
	
\end{figure}

On a number of not very large datasets, the time taken to calculate meta-features in the previous section are already non-negligible, as shown in Figure~\ref{fig:metafeature_calculation}. Each dot represents one midsize OpenML dataset.

\newpage
\section{Comparison of experiment design with different constraints}
\label{sec:comparison_of_different_ED}
\begin{figure}[H]
	\centering
	\includegraphics[width=\linewidth]{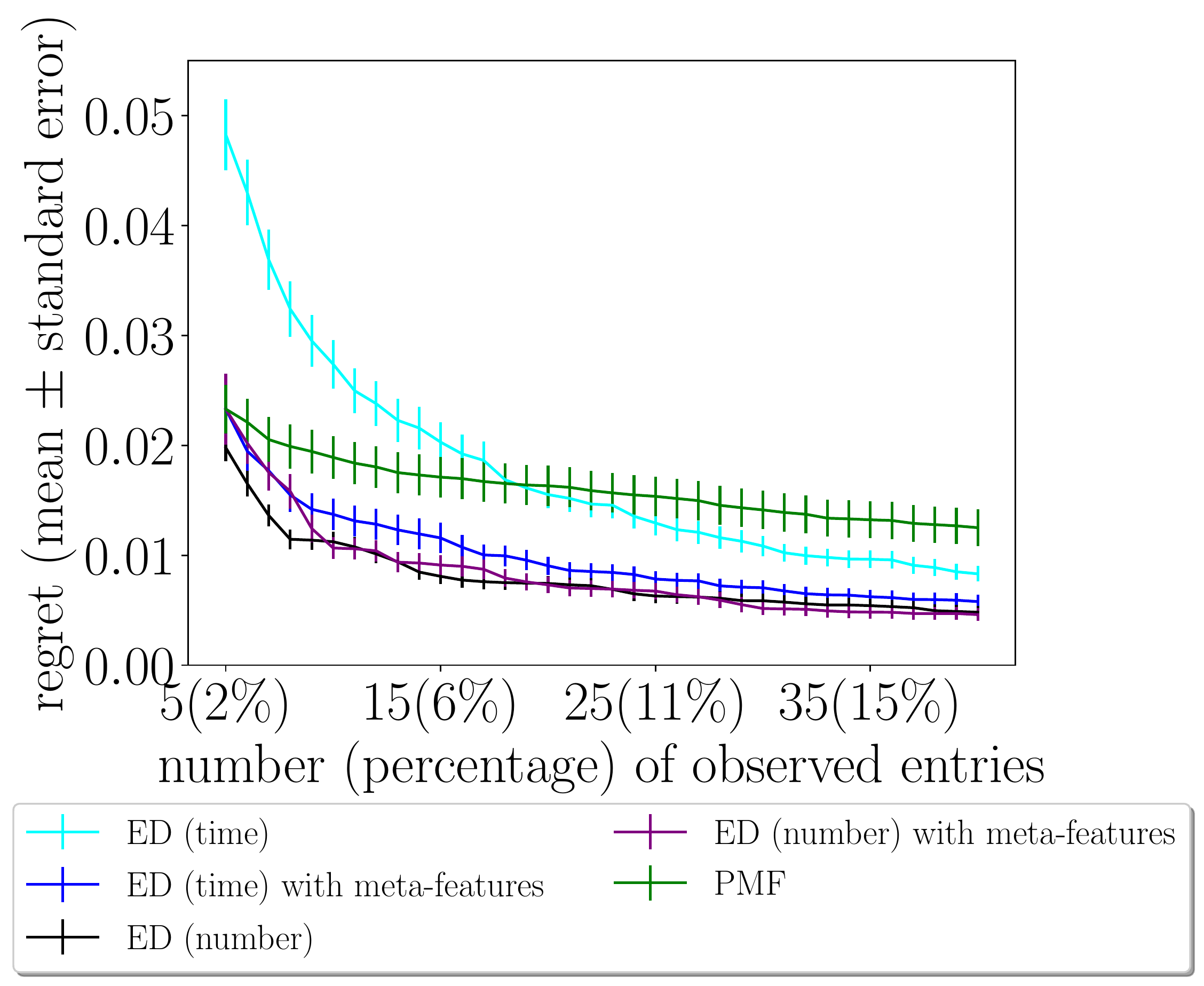}
	\caption{Comparison of different versions of ED with PMF. "ED (time)" denotes ED with runtime constraint, with time limit set to be 10\% of the total runtime of all available models; "ED (number)" denotes ED with the number of entries constrained.}
	\label{fig:comparison_of_different_ED}
\end{figure}

In Section~\ref{section:numerics_of_comparison_with_pmf}, our experiments compare QR and PMF to 
a variant of experiment design (ED) with a constraint on the number of observed entries,
since QR and PMF admit a similar constraint.
Figure~\ref{fig:comparison_of_different_ED} shows that the regret of ED with a runtime constraint (Equation~\ref{equation:ED_time})
is not too much larger.

\end{document}